\documentclass[11pt]{article}

\usepackage[final]{acl}

\usepackage{times}
\usepackage{latexsym}

\usepackage[T1]{fontenc}

\usepackage[utf8]{inputenc}

\usepackage{microtype}
\usepackage{inconsolata}
\usepackage{graphicx}
\usepackage{xcolor}

\usepackage{url}
\usepackage{amsmath}
\usepackage{amsfonts}
\usepackage{subcaption}
\usepackage{tcolorbox}
\tcbuselibrary{skins,breakable}
\usepackage{listings}
\usepackage{booktabs}
\usepackage{tabularx}
\usepackage{multirow}
\usepackage[normalem]{ulem}

\newcommand{\best}[1]{\textbf{#1}}
\newcommand{\second}[1]{\uline{#1}}
\tcbset{
  findingbox/.style={
    colback=gray!5,             
    colframe=black!30,          
    coltitle=white,             
    colbacktitle=black!60,      
    fonttitle=\normalsize\bfseries,      
    boxrule=0.3pt,              
    arc=2pt,                    
    left=6pt, right=6pt, top=2pt, bottom=2pt, 
    enhanced,
    boxed title style={size=small, colframe=black!60, colback=black!60, sharp corners, top=1pt, bottom=1pt, left=2pt, right=2pt},
  }
}

\title{When Does Reasoning Help in Machine Translation? A Hierarchical Analysis of LRM Reasoning Traces}

\author{
Yuxiang Liu$^{1}$\thanks{Work conducted during an internship at Google.}\thanks{Corresponding author.} \quad
Jiaming Luo$^{2}$ \quad
Eleftheria Briakou$^{2}$ \quad
Colin Cherry$^{2}$ \\
$^{1}$University of Illinois at Urbana-Champaign
\quad
$^{2}$Google DeepMind \\
\texttt{yuxiang@illinois.edu}
\quad
\texttt{\{jmluo,ebriakou,colincherry\}@google.com}
}

\begin{document}
\maketitle

\begin{abstract}


Large Reasoning Models increasingly use intermediate traces for machine translation, but it remains unclear when such reasoning helps or hurts.
We analyze reasoning traces across models, languages, domains, and datasets, focusing on reasoning language, length, and structure.
We find that the best reasoning language is model-specific, reasoning length has a non-monotonic relationship with quality, and traces exhibit recurring functional patterns.
To uncover these patterns, we introduce \textbf{Hierarchical Meta-Summarization (HMS)}, a scalable framework that induces coarse- and fine-grained reasoning structures without predefined taxonomies.
HMS reveals a shared organization--understanding/planning, translating/drafting, and refining/verifying--alongside domain-specific variation.
Our results suggest that MT reasoning should be controlled in a model-aware, length-aware, and pattern-aware manner rather than uniformly encouraged.


\end{abstract}

\section{Introduction}

Large Language Models have improved machine translation (MT), especially in long-form, low-resource, and culturally nuanced settings~\citep{briakou2024translating,zebaze2025context,anik2025preserving}.
Large Reasoning Models (LRMs) extend this paradigm by generating reasoning traces before translation~\citep{ye2025well,chen2025evaluating,liu2025new}.
However, this added reasoning brings new challenges: models may reason in ineffective languages, produce unnecessarily long traces, or emphasize unhelpful steps.
MT reasoning is therefore not only a capability to enable, but also a control problem: deciding \emph{which language} to reason in, \emph{how much} reasoning to produce, and \emph{which steps} to emphasize.

Prior work on reasoning-enabled MT mainly treats reasoning as a means to improve translation quality~\citep{wang2024drt,nguyen2025reasoning,he2025r1}, leaving the traces themselves underexplored.
As a result, it remains unclear which trace properties are useful, when longer traces add unnecessary computation, and how these effects vary across models, language pairs, and domains.
We therefore analyze reasoning through a resource-allocation perspective, focusing on variation in trace language, length, and structural patterns.

Existing reasoning trace analyses are not well suited to MT.
Most focus on mathematical or logical tasks with discrete answers~\citep{yong2025crosslingual,wang2025language}; even studies of cross-lingual reasoning imbalance~\citep{bajpai2025multilingual, qi2025models, park2025cross} and overthinking~\citep{ballon2025relationship, chen2024not, su2025between} do not directly explain their relationship to translation quality.
Moreover, structural analyses often rely on fixed, flat taxonomies~\citep{yang2025understanding, hu2025distillation, marjanovic2025deepseek, xiao2025limopro} or localized trace-level structures~\citep{xiong2025mapping}, limiting their ability to capture domain-specific MT reasoning patterns.
We therefore need a scalable method for inducing reasoning functions directly from MT traces.

We present a systematic empirical study of reasoning traces in LRM-based MT, centered on three questions:
(i) \textbf{reasoning language}: whether models follow the instructed language and how this behavior relates to translation quality;
(ii) \textbf{reasoning length}: whether longer traces are associated with better translations or instead indicate overthinking;
(iii) \textbf{reasoning structure}: how models allocate reasoning across functional stages such as planning, drafting, and verification.
To support the structural analysis, we introduce \textbf{Hierarchical Meta-Summarization (HMS)}, a scalable, data-driven framework that abstracts raw traces into atomic functional steps and recursively induces coarse- and fine-grained reasoning patterns without predefined taxonomies or manual annotation.

Our analysis yields three main findings.
First, the most effective reasoning language is model-specific and tends to align with the language a model follows most reliably.
Second, reasoning length has a non-monotonic relationship with translation quality: moderate traces can be competitive or beneficial, whereas longer traces often correlate with degradation.
Third, HMS reveals a recurring high-level structure in general-purpose LRMs--\emph{Understanding/Planning}, \emph{Translating/Drafting}, and \emph{Refining/Verifying}--with domain-specific fine-grained allocation and pattern-quality associations.
Together, these results suggest that effective LRM-based MT requires model-aware, length-aware, and pattern-aware control rather than simply encouraging more reasoning.

Our contributions are as follows:

\noindent (1) We systematically analyze reasoning language and length in LRM-based MT, showing their importance for translation quality and efficiency.

\noindent (2) We propose Hierarchical Meta-Summarization, a scalable framework for inducing hierarchical functional patterns from reasoning traces.

\noindent (3) We show that general-purpose LRMs share high-level reasoning structure but vary in domain-specific allocation and pattern-quality associations.


\section{Related Work}
\label{sec:related}

\subsection{LLM-based Machine Translation}

Recent work on LLM-based machine translation (MT) has studied model behavior and methods for improving translation quality.
One line of work examines mechanisms such as attention heads and MLPs~\cite{zhang2025exploring}, as well as inference-time choices such as prompting strategies~\cite{nguyen2025reasoning}.
Another improves MT through reasoning, including step-level control~\cite{briakou2024translating} and curated reasoning data~\cite{wang2024drt,he2025r1}.
Large Reasoning Models (LRMs), which generate intermediate reasoning traces, outperform standard LLMs on MT benchmarks, especially for complex and long-context translation tasks~\cite{ye2025well,liu2025new,chen2025evaluating}.
\textbf{However, prior work uses reasoning to improve translation performance but leaves the resulting traces largely unanalyzed. We provide a systematic empirical analysis of reasoning traces in LRM-based MT.}

\subsection{Large Reasoning Models (LRMs)}

LRMs extend LLMs with explicit reasoning and perform well on complex tasks, but their behavior in MT--a cross-lingual and efficiency-sensitive setting--remains underexplored.
First, reasoning abilities learned in high-resource languages may transfer poorly to other languages.
Models can mix languages or revert to a dominant pretraining language~\cite{bajpai2025multilingual,qi2025models,park2025cross}, often performing better in high-resource reasoning languages~\cite{wang2025polymath}.
Although such gaps have been studied in mathematics and logic~\cite{yong2025crosslingual,wang2025language}, their impact on MT remains unclear.
Second, LRMs often generate overly long reasoning traces with diminishing, or even negative, returns, a phenomenon known as overthinking~\cite{ballon2025relationship,chen2024not}.
This inefficiency is commonly attributed to difficulty miscalibration, where models overthink simple inputs and underthink harder ones~\cite{shen2025dast,su2025between}.
Yet its effect on translation quality remains insufficiently studied.
\textbf{Together, these gaps motivate our analysis of how reasoning language and reasoning length affect translation quality and efficiency in LRM-based MT.}

\subsection{Reasoning Pattern Analysis}

Reasoning patterns reveal how LRMs decompose problems, revise decisions, and validate outputs.
Prior work typically relies on fixed or manually defined patterns, including token-level markers of self-reflection or correction~\cite{yang2025understanding,hu2025distillation,liu2025exploring}, predefined stage decompositions~\cite{marjanovic2025deepseek}, and functional distinctions between core reasoning and meta-cognitive behaviors, such as backtracking and verification~\cite{xiao2025limopro,stechly2025beyond}.
Although informative, these approaches depend on predefined taxonomies or manual annotation, which limits scalability and can obscure task- and domain-specific variation.
\textbf{In contrast, we propose an automatic, hierarchical framework that induces reasoning patterns from data without predefined categories.}

\section{Preliminary Analysis}
\label{sec:preliminary}

This section presents a preliminary trace-level analysis of reasoning language and length.

\begin{table*}[tp!]
    \centering
    \begin{tabular}{lcccccccccc}
        \toprule
        \textbf{Inst.$\rightarrow$Reason}
        & \multicolumn{5}{c}{\textbf{WMT}}
        & \multicolumn{3}{c}{\textbf{CMT}}
        & \textbf{DRT} & \textbf{Mean} \\
        \cmidrule(lr){2-6} \cmidrule(lr){7-9} \cmidrule(lr){10-10}
        & \textbf{en-de} & \textbf{en-es} & \textbf{en-ja} & \textbf{en-ru} & \textbf{en-zh}
        & \textbf{en-es} & \textbf{en-fr} & \textbf{en-zh}
        & \textbf{en-zh} & \\
        \midrule
        en$\rightarrow$en (\%)
        & \second{78.6} & \second{93.3} & 29.4 & \second{66.8} & \second{64.2}
        & \second{92.5} & \second{92.9} & \second{57.0}
        & \second{63.8} & \second{71.0} \\
        zh$\rightarrow$zh (\%)
        & \best{97.2} & \best{98.3} & \best{92.1} & \best{94.8} & \best{99.2}
        & \best{97.5} & \best{97.8} & \best{99.8}
        & \best{99.0} & \best{97.3} \\
        tgt$\rightarrow$tgt (\%)
        & 27.7 & 7.2 & \second{69.3} & 55.0 & -- & 4.6 & 7.3 & -- & -- & 28.5 \\
        \bottomrule
    \end{tabular}%
    \caption{Reasoning-Language Instruction-Adherence (RLIA) rates for Qwen-32B. The model follows Chinese reasoning-language instructions most reliably, while target-language instructions yield much lower adherence.}
    \label{tab:qwen32b-instruction-adherence}
\end{table*}

\subsection{Experiment Setup}

To systematically analyze reasoning in machine translation (MT), we evaluate multilingual Large Reasoning Models (LRMs) on diverse translation tasks. Each model is instructed to generate a reasoning trace before producing its translation, yielding a large corpus of traces for analysis.

\subsubsection{Translation Datasets}

We use three datasets that present complementary translation challenges: WMT24++~\cite{deutsch2025wmt24++}, CultureMT~\cite{yao2023benchmarking}, and DRT-Literature~\cite{wang2024drt}.

\paragraph{WMT24++ (WMT)} A large-scale dataset covering four domains: literary, news, social, and speech. We analyze five language pairs: English-German (en-de), English-Spanish (en-es), English-Japanese (en-ja), English-Russian (en-ru), and English-Chinese (en-zh).

\paragraph{CultureMT (CMT)} A culturally focused dataset targeting culture-specific items. We select three language pairs for reasoning over culturally grounded content: English-Spanish (en-es), English-French (en-fr), and English-Chinese (en-zh).

\paragraph{DRT-Literature (DRT)} An English-Chinese (en-zh) literary dataset designed for deep translation. Although it overlaps with WMT24++' literary domain, DRT enables a focused analysis of long, complex translations that require sustained reasoning.

\subsubsection{LLM Setup}\label{sec:LLMSetup}

\paragraph{Model Selection} Prior work shows that LRMs outperform non-reasoning counterparts on MT, especially for complex, long-context, and reasoning-intensive inputs~\cite{ye2025well,liu2025new,chen2025evaluating}. We therefore study how reasoning operates within LRMs rather than re-establishing its benefits. We evaluate six representative models: DeepSeek-R1-Distill-Qwen-14B (\textbf{Qwen-14B}), DeepSeek-R1-Distill-Qwen-32B (\textbf{Qwen-32B}), \textbf{gpt-oss-20B}, DeepSeek-R1-Distill-Llama-8B (\textbf{Llama-8B}), gemma-4-E4B-it (\textbf{Gemma-4-E4B}), and the MT-specific \textbf{DRT-14B}~\cite{wang2024drt}.

\paragraph{Prompting Strategy} We prompt each model to reason in a specified language before producing its translation. The reasoning language is English, Chinese, or the target language for non-Chinese targets. We use the following prompt template:
\begin{tcolorbox}[findingbox, title=Translation Prompt]
Please always think in \textbf{\{reasoning language\}}. Translate the following text from \textbf{\{source language\}} to \textbf{\{target language\}}:
\begin{verbatim}
{source text here}
\end{verbatim}
\end{tcolorbox}

\paragraph{Sampling Parameters} We use a temperature of $0.6$ and top-$p$ of $0.95$ to promote output diversity. For each source text, we generate $16$ samples to capture diverse reasoning traces and reduce evaluation variance.

\subsubsection{Translation Metrics}

We evaluate translation quality using three learned metrics: COMET-22 (\textbf{COMET})~\cite{rei2022comet}, which predicts human judgments of MT quality; MetricX (\textbf{MX}), a reference-based MetricX variant; and MetricX-QE (\textbf{MX-QE})~\cite{juraska2024metricx}, a reference-free variant. MX uses the reference translation, whereas MX-QE estimates quality from only the source and hypothesis. Because MetricX is an error metric, where lower is better, we report \textit{normalized} quality-oriented scores, \textbf{Norm.MX} and \textbf{Norm.MX-QE}, computed as $1-\frac{s}{25}$, where $s$ is the raw MetricX or MetricX-QE score. Higher normalized MetricX values therefore indicate better translation quality, consistent with COMET.
\subsection{Reasoning Language Analysis}\label{sec:language}

\begin{table*}[tp!]
    \centering
    \footnotesize
    \begin{tabular}{llcccccccccc}
        \toprule
        \textbf{Metric} & \textbf{Inst.}
        & \multicolumn{5}{c}{\textbf{WMT}}
        & \multicolumn{3}{c}{\textbf{CMT}}
        & \textbf{DRT}
        & \textbf{Mean} \\
        \cmidrule(lr){3-7} \cmidrule(lr){8-10} \cmidrule(lr){11-11}
        &
        & \textbf{en-de} & \textbf{en-es} & \textbf{en-ja} & \textbf{en-ru} & \textbf{en-zh}
        & \textbf{en-es} & \textbf{en-fr} & \textbf{en-zh}
        & \textbf{en-zh}
        & \\
        \midrule
        \multirow{2}{*}{COMET$\uparrow$}
        & en & \textbf{0.7613} & 0.7774 & 0.8186 & 0.7631 & 0.8297 & 0.8063 & \textbf{0.7676} & 0.8361 & 0.7742 & 0.7927 \\
        & zh & 0.7442 & \textbf{0.7976} & \textbf{0.8267} & \textbf{0.7662} & \textbf{0.8363} & \textbf{0.8135} & 0.7636 & \textbf{0.8404} & \textbf{0.7810} & \textbf{0.7966} \\
        \midrule
        \multirow{2}{*}{Norm.MX$\uparrow$}
        & en & 0.8640 & 0.8293 & 0.7685 & 0.7728 & 0.8825 & 0.8456 & 0.8219 & 0.8634 & 0.8400 & 0.8320 \\
        & zh & \textbf{0.8710} & \textbf{0.8526} & \textbf{0.7842} & \textbf{0.7807} & \textbf{0.8894} & \textbf{0.8678} & \textbf{0.8296} & \textbf{0.8715} & \textbf{0.8502} & \textbf{0.8441} \\
        \midrule
        \multirow{2}{*}{Norm.MX-QE$\uparrow$}
        & en & 0.8766 & 0.8358 & 0.7845 & 0.7977 & 0.8884 & 0.8605 & 0.8676 & 0.8769 & 0.8361 & 0.8471 \\
        & zh & \textbf{0.8852} & \textbf{0.8620} & \textbf{0.8013} & \textbf{0.8059} & \textbf{0.8960} & \textbf{0.8839} & \textbf{0.8764} & \textbf{0.8855} & \textbf{0.8472} & \textbf{0.8604} \\
        \bottomrule
    \end{tabular}
    \caption{Translation quality for Qwen-32B by instructed reasoning language. Chinese reasoning yields higher mean scores across all three metrics, especially on Norm.MX and Norm.MX-QE.}
    \label{tab:qwen32b-quality-by-inst-lang}
\end{table*}

We first examine how the instructed reasoning language affects adherence and translation quality.
\textbf{Reasoning-Language Instruction-Adherence (RLIA)} measures the proportion of traces detected as the instructed language $l$\footnote{We use an in-house language identification detector.}:
\begin{equation}
\mathrm{RLIA}(l) = \frac{1}{N} \sum_{i=1}^{N}
\mathbf{1}\!\left[\operatorname{LangID}(r_i)=l\right],
\end{equation}
where $r_i$ is the reasoning trace for output $i$.

\begin{table}[tp!]
    \centering
    \begin{tabular}{lcc}
        \toprule
        \textbf{Model} & \textbf{en$\rightarrow$en (\%)} & \textbf{zh$\rightarrow$zh (\%)} \\
        \midrule
        Qwen-14B      & 71.6 & \textbf{97.6} \\
        Qwen-32B      & 64.2 & \textbf{99.2} \\
        gpt-oss-20B   & \textbf{98.3} & 2.4 \\
        Llama-8B      & 77.8 & \textbf{99.9} \\
        Gemma-4-E4B   & \textbf{95.3} & 2.1 \\
        \bottomrule
    \end{tabular}
    \caption{RLIA on WMT (en-zh). Qwen and Llama favor Chinese reasoning instructions, whereas gpt-oss-20B and Gemma-4-E4B favor English.}
    \label{tab:wmt24pp-enzh-instruction-adherence}
\end{table}

\begin{table}[tp!]
    \centering
    \small
    \begin{tabular}{@{}lcccc@{}}
        \toprule
        \textbf{Model} & \textbf{Inst.} & \textbf{COMET$\uparrow$} & \textbf{MX$\uparrow$} & \textbf{MX-QE$\uparrow$} \\
        \midrule
        \multirow{2}{*}{Qwen-14B}
        & en & 0.8227 & 0.8644 & 0.8730 \\
        & zh & \textbf{0.8285} & \textbf{0.8749} & \textbf{0.8838} \\
        \midrule
        \multirow{2}{*}{Qwen-32B}
        & en & 0.8297 & 0.8825 & 0.8884 \\
        & zh & \textbf{0.8363} & \textbf{0.8894} & \textbf{0.8960} \\
        \midrule
        \multirow{2}{*}{gpt-oss-20B}
        & en & \textbf{0.8352} & \textbf{0.8837} & \textbf{0.8935} \\
        & zh & 0.8326 & 0.8791 & 0.8886 \\
        \midrule
        \multirow{2}{*}{Llama-8B}
        & en & 0.7939 & 0.8374 & 0.8510 \\
        & zh & \textbf{0.8035} & \textbf{0.8468} & \textbf{0.8612} \\
        \midrule
        \multirow{2}{*}{Gemma-4-E4B}
        & en & \textbf{0.8413} & \textbf{0.9106} & \textbf{0.9128} \\
        & zh & 0.6525 & 0.8480 & 0.8525 \\
        \bottomrule
    \end{tabular}
    \caption{Translation quality on WMT (en-zh). Chinese reasoning instructions perform better for Qwen and Llama, whereas English reasoning instructions perform better for gpt-oss-20B and Gemma-4-E4B.}
    \label{tab:wmt24pp-enzh-quality-by-inst-lang}
\end{table}

\paragraph{RLIA is Model-Specific}
Qwen-32B adheres most strongly to Chinese reasoning instructions across dataset-language pairs, with 97.3\% mean RLIA, compared with 71.0\% for English and 28.5\% for target-language reasoning (Table~\ref{tab:qwen32b-instruction-adherence}).\footnote{Due to limited coherent reasoning in other target languages, we restrict subsequent analyses to English and Chinese.} Qwen-14B shows the same trend (Table~\ref{tab:qwen14b-instruction-adherence}; Appendix~\ref{app:qwen14b-reasoning-language}).
However, this preference is model-specific: on WMT (en-zh), Qwen-14B, Qwen-32B, and Llama-8B adhere more strongly to Chinese instructions, whereas gpt-oss-20B and Gemma-4-E4B favor English (Table~\ref{tab:wmt24pp-enzh-instruction-adherence}).
Overall, RLIA reveals \textbf{model-specific reasoning-language preferences}, not a universal advantage for either language. Appendix~\ref{app:reasoning-language-control} compares explicit language control with an unconstrained baseline.

\begin{figure*}[tp!]
    \centering
    \includegraphics[width=0.95\textwidth]{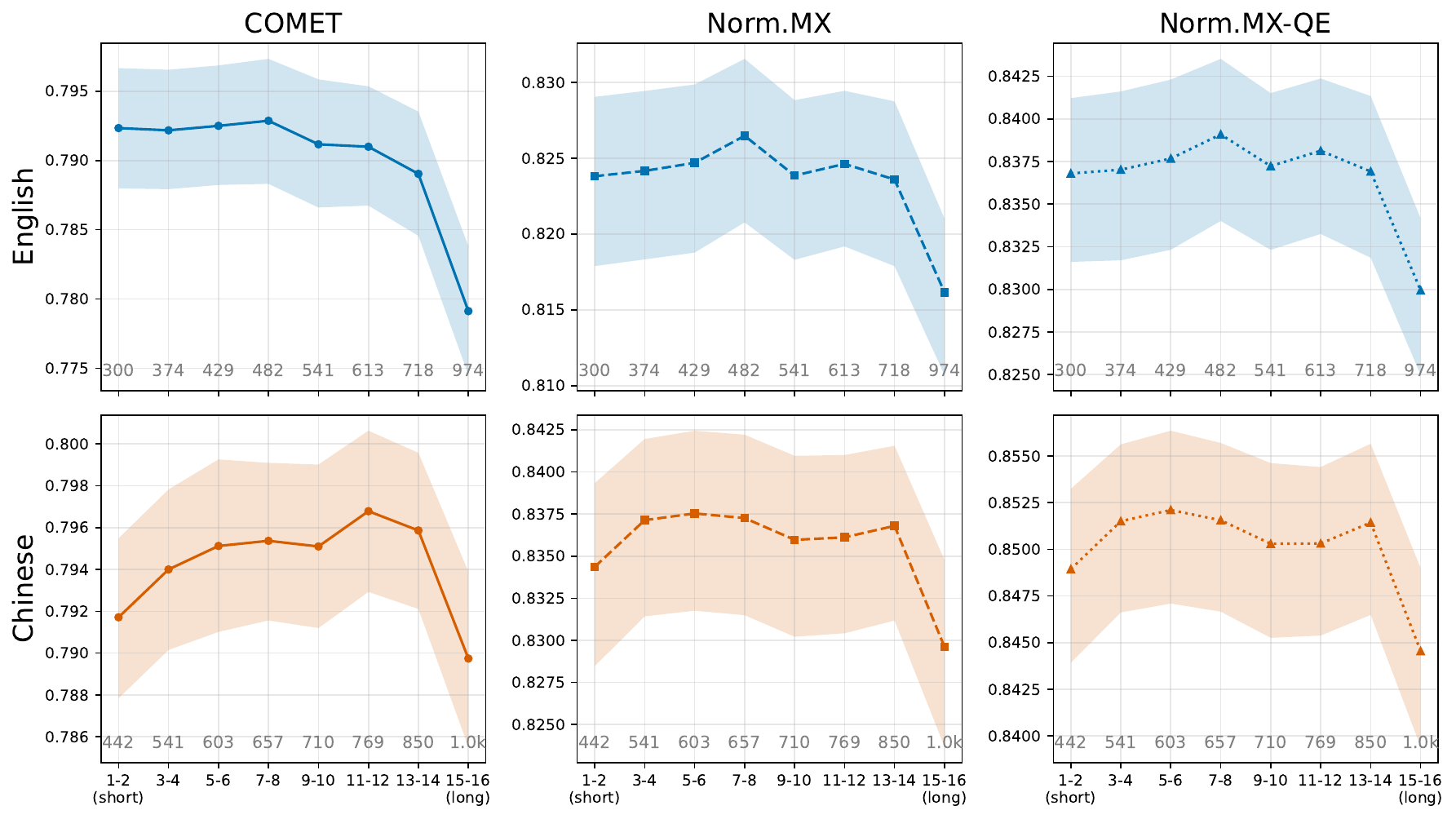}
    \caption{Translation quality by ranked reasoning trace length for Qwen-32B on WMT.}
    \label{fig:wmt24-qwen32B-reasoning-length}
\end{figure*}

\paragraph{Cognitive Consistency in Translation Quality}
Translation quality follows the same model-specific pattern as RLIA.
For Qwen-32B, Chinese reasoning yields the highest mean score across all metrics (Table~\ref{tab:qwen32b-quality-by-inst-lang}). Qwen-14B shows the same trend, particularly on the Norm.MX-based metrics (Table~\ref{tab:qwen14b-quality-by-inst-lang}, Appendix~\ref{app:qwen14b-reasoning-language}).
However, Chinese is not universally optimal. On WMT (en-zh), Chinese instructions yield better results for Qwen-14B, Qwen-32B, and Llama-8B, whereas English instructions yield better results for gpt-oss-20B and Gemma-4-E4B (Table~\ref{tab:wmt24pp-enzh-quality-by-inst-lang}).
Thus, translation quality aligns with RLIA: \textbf{models tend to translate better when reasoning in the language they follow most reliably}.
We refer to this model-specific alignment as \emph{Cognitive Consistency}.
This association does not imply that higher RLIA causes better translation quality. Instead, a model's underlying language proficiency or training exposure may lead to both higher RLIA and better translation quality.
Reasoning language should therefore be selected for each model. RLIA can serve as an inexpensive diagnostic for identifying promising reasoning languages, but not as a causal predictor of translation quality.
\subsection{Reasoning Length Analysis}\label{sec:ReasoningEfficiency}

LRMs can produce reasoning traces of widely varying lengths for the same source, raising the risk of \emph{overthinking}: expending additional computation without improving output quality~\cite{chen2024not,su2025between,shen2025dast}.

To study this effect, we sample 16 traces per source (Section~\ref{sec:LLMSetup}), rank them by length, group adjacent ranks into eight bins from \(r_{1\text{-}2}\) to \(r_{15\text{-}16}\), and evaluate translation quality for each bin. This within-sentence comparison controls for source difficulty while remaining observational.
We use a 10,000-replicate, domain-stratified document-cluster bootstrap. Within each domain, we resample documents with replacement, retaining all segments and generations; in each replicate, we re-rank the 16 generations per source and condition, recompute the bin-level mean scores, and average across language pairs within each dataset.

Figure~\ref{fig:wmt24-qwen32B-reasoning-length} shows the main result for Qwen-32B on WMT, with adjacent ranks grouped into eight bins from \(r_{1\text{-}2}\) to \(r_{15\text{-}16}\) (see Appendix~\ref{app:reasoning-length-stats} for length statistics).
Translation quality is generally stable or improves modestly from short to intermediate reasoning lengths, but tends to deteriorate at the greatest lengths; the magnitude of this effect varies across datasets, reasoning languages, and metrics.
This non-monotonic trend also appears across the additional models and datasets reported in Appendix~\ref{app:reasoning-length-quality-curves}, where the longest traces often yield limited gains or degrade quality.

A controlled experiment in Appendix~\ref{app:reasoning-length-control} further shows that explicitly increasing gpt-oss-20B's reasoning effort nearly doubles its average trace length, yet produces only small, metric-dependent quality changes.
Together, the observational and controlled results suggest a useful middle range, beyond which additional tokens may reflect inefficient or unstable generation rather than better translation. Reasoning length alone is therefore an unreliable indicator of MT quality, supporting adaptive rather than uniformly increased reasoning.

\section{Hierarchical Meta-Summarization}
\label{sec:HMS}

\begin{figure*}[htbp]
    \centering
    \includegraphics[width=\textwidth]{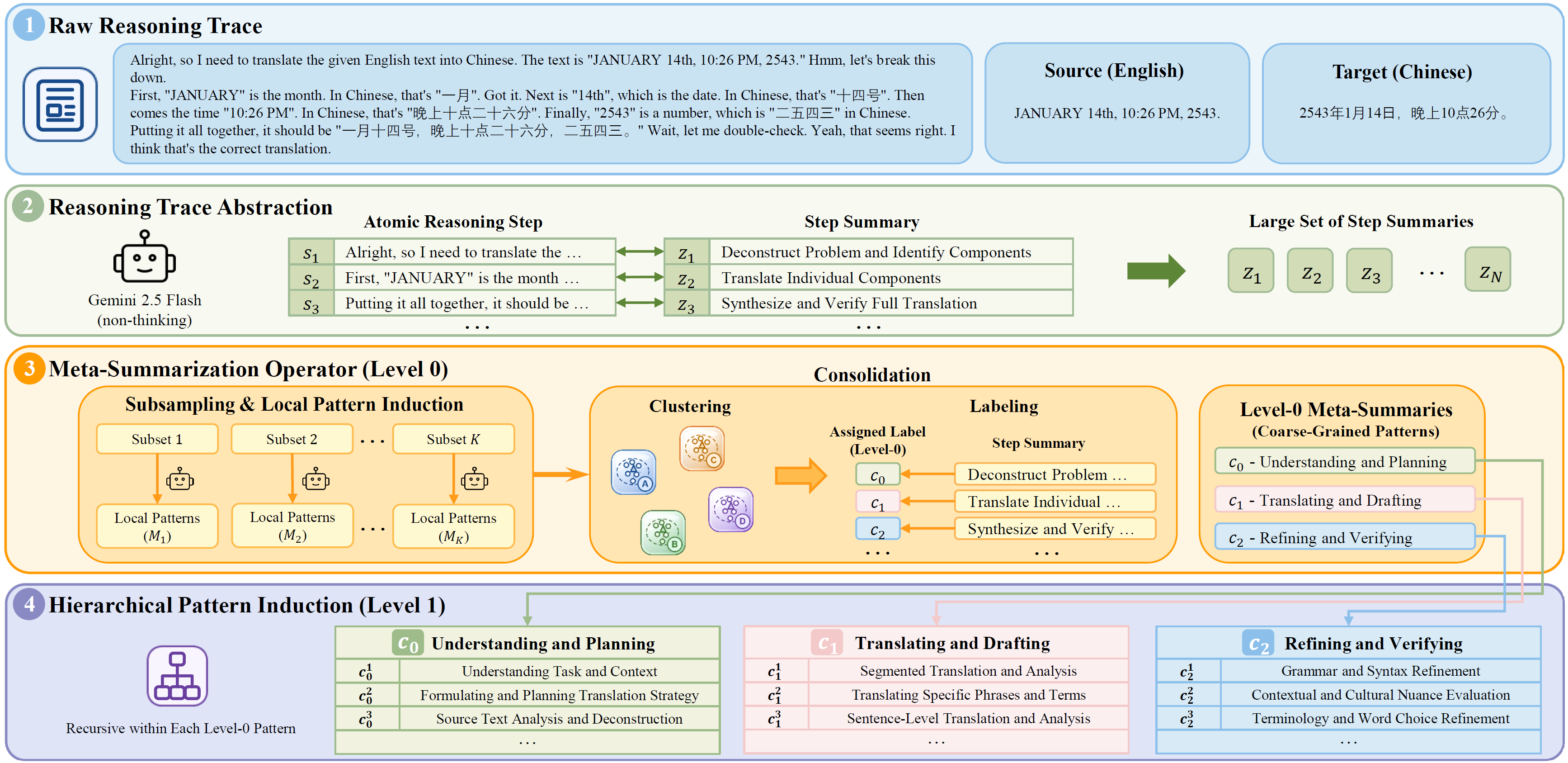}
    \caption{Hierarchical Meta-Summarization (HMS). HMS is a data-driven framework for structural reasoning pattern analysis that abstracts raw reasoning traces into atomic steps and recursively applies meta-summarization to derive coarse Level-0 patterns and finer Level-1 sub-patterns.}
    \label{fig:HMSFramework}
\end{figure*}

To move beyond trace-level analyses of reasoning language and length, we characterize structural patterns in reasoning traces. Prior work focuses on coarse token-level distinctions~\cite{yang2025understanding, hu2025distillation}, uses fixed categories~\cite{marjanovic2025deepseek, xiao2025limopro}, or analyzes individual traces without capturing shared patterns~\cite{xiong2025mapping}. We propose \textbf{Hierarchical Meta-Summarization} (HMS; Figure~\ref{fig:HMSFramework}), a recursive framework to derive multi-level reasoning patterns from large trace collections. HMS abstracts traces into atomic reasoning steps and recursively meta-summarizes them to derive dataset-level patterns (Level 0) and category-specific sub-patterns (Level 1).\footnote{We provide an HMS cost breakdown in Appendix~\ref{app:hms-cost-breakdown} and human validation of HMS summaries and labels in Appendix~\ref{app:hms-validation}.}

\subsection{Reasoning Trace Abstraction}

HMS first normalizes raw reasoning traces into atomic functional units. Given a trace set $\mathbb{R}$, each trace is segmented into coherent reasoning steps and abstracted into concise functional summaries, which standardizes step granularity and reduces textual complexity for downstream analysis.
We use Gemini 2.5 Flash~\cite{comanici2025gemini} to jointly perform segmentation and abstraction in a single pass\footnote{We use the non-thinking mode throughout.}. For each trace $r \in \mathbb{R}$, the model acts as a \textit{Reasoning Step Abstractor} (prompt in Appendix~\ref{prompt:segment&summarize}) and produces step-summary pairs:
\[
r \xrightarrow{\text{Gemini}} \{(s_1, z_1), (s_2, z_2), \ldots, (s_n, z_n)\},
\]
where $s_i$ is a reasoning step and $z_i$ is its functional summary. These summaries provide a normalized trace representation for meta-summarization.

\subsection{Meta-Summarization Operator}
\label{sec:meta-summarization-operator}

The abstraction stage yields a large set of step-level summaries, $\mathbb{Z} = \{z_1, \ldots, z_N\}$, where $N$ is the number of atomic reasoning steps across all traces in $\mathbb{R}$. Meta-summarization aggregates these summaries into a compact set of high-level reasoning patterns, or \textit{Meta-Summaries}, $\mathbf{M}$.
In practice, $\mathbb{Z}$ is too large for direct clustering or single-pass LLM analysis\footnote{For WMT24, each trace averages around 450 tokens (approximately 7 steps), yielding nearly 860k step-level summaries after abstraction.}. To induce stable global patterns at this scale, we adopt a subsampling-based strategy: each subset provides a partial view of the reasoning structure, and the resulting redundant meta-summaries are consolidated into a unified set.

\subsubsection{Subsampling-Based Pattern Induction}

To balance corpus coverage and inference stability, we sample $K$ independent subsets without replacement: $Z_k \subset \mathbb{Z}$, where $|Z_k| = N_s$ and $k = 1, \ldots, K$\footnote{We use $N_s=500$; increasing $N_s$ to 1000 led to more degenerate outputs under heuristic rule checking. We set $K=128$ in our experiments.}.
For each subset, the model identifies emergent functional clusters and summarizes them as high-level reasoning patterns (prompt in Appendix~\ref{prompt:meta-summarize}): $Z_k \xrightarrow{\text{Gemini}} \mathbf{M}_k = \{m_{k,1}, \ldots, m_{k,J_k}\}$, where each $m_{k,j}$ describes a shared reasoning function. Aggregating these outputs yields the pooled set $\mathbf{M}_{\text{total}} = \bigcup_{k=1}^{K} \mathbf{M}_k$.

\subsubsection{Consolidation}

The pooled set $\mathbf{M}_{\text{total}}$ contains substantial redundancy and lexical variation. To derive stable corpus-level reasoning patterns, we apply a two-stage consolidation procedure.

\paragraph{Clustering.} 
Each candidate $m \in \mathbf{M}_{\text{total}}$ is encoded using a pretrained sentence embedding model\footnote{\url{https://ai.google.dev/gemini-api/docs/embeddings}}. We then use K-means to cluster the candidates into $L$ consolidated categories.\footnote{Details on the \textit{data-guided} procedure for selecting $L$ are provided in Appendix~\ref{app:kmeans-details}.} The representative descriptions of these categories constitute the final \textit{Meta-Summaries}, $\mathbf{M}_{\text{final}}$.

\paragraph{Labeling.}
We assign each step-level summary $z_i \in \mathbb{Z}$ a reasoning pattern label via LLM-based classification (Appendix~\ref{prompt:label}). Given the original step $s_i$, its summary $z_i$, and $\mathbf{M}_{\text{final}}$, the model predicts the label $l_i$: $s_i, \; z_i, \; \mathbf{M}_{\text{final}} \xrightarrow{\text{Gemini}} l_i$.

\subsection{Hierarchical Pattern Induction}

HMS constructs a reasoning hierarchy by applying the meta-summarization operator (Section~\ref{sec:meta-summarization-operator}) at multiple scopes. It first induces coarse-grained dataset-level patterns, then recurses within each pattern to uncover finer-grained sub-structures.

\paragraph{Level-0 Meta-Summarization.}
The operator is first applied to all step-level summaries $\mathbb{Z}$, inducing high-level reasoning patterns, referred to as Level-0 categories and denoted by $\mathbf{M}_{\text{final}}$.
Each summary $z_i \in \mathbb{Z}$ is assigned a Level-0 label $l_i \in \mathbf{M}_{\text{final}}$ for its primary functional role.

\paragraph{Level-1 Meta-Summarization.}
To capture finer-grained structure, we apply the same operator within each Level-0 category.
Specifically, $\mathbb{Z}$ is partitioned by Level-0 label into subsets $\mathbb{Z}_m = \{ z_i \in \mathbb{Z} \mid l_i = m \}$ for $m \in \mathbf{M}_{\text{final}}$.
Meta-summarization is then applied to each $\mathbb{Z}_m$ to induce category-specific sub-patterns $\mathbf{M}_m^{(1)}$.
Each summary step $z_i$ receives a Level-1 label $l_i^{(1)} \in \mathbf{M}_{l_i}^{(1)}$, yielding the hierarchical label pair $(l_i, l_i^{(1)})$.

\section{Hierarchical Pattern Analysis}\label{sec:analysis}

Using the reasoning patterns induced by HMS in Section~\ref{sec:HMS}, we examine how they are instantiated in practice. Specifically, we study how reasoning effort is allocated across patterns and how stage effectiveness relates to translation quality.

\subsection{Domain-General Reasoning Patterns}

\begin{table}[tp!]
\centering
\scriptsize
\setlength{\tabcolsep}{2pt}
\renewcommand{\arraystretch}{1.05}
\resizebox{\linewidth}{!}{%
\begin{tabular}{p{0.20\linewidth}p{0.62\linewidth}p{0.15\linewidth}}
\toprule
\textbf{Model} & \textbf{Induced HMS cluster} & \textbf{Function} \\
\midrule
\multirow{3}{*}{Gemma-4-E4B}
& Analyzing Source Text & U/P \\
& Drafting Initial Translation & T/D \\
& Refining and Polishing Translation & R/V \\
\midrule
\multirow{3}{*}{Llama-8B}
& Initial Understanding and Planning & U/P \\
& Executing Translation & T/D \\
& Refining and Verifying & R/V \\
\midrule
\multirow{3}{*}{gpt-oss-20B}
& Initial Analysis and Understanding & U/P \\
& Drafting and Refining Translation & T/D \& R/V \\
& Finalization and Output & R/V \\
\midrule
\multirow{3}{*}{DRT-14B}
& Initial Translation \& Problem Identification & U/P \& T/D \\
& Refinement \& Revision & R/V \\
& Evaluation \& Critique & R/V \\
\bottomrule
\end{tabular}
}
\caption{Cross-model mapping of induced HMS clusters to shared functions. General-purpose LRMs largely recover the three-stage structure, whereas MT-specialized DRT-14B emphasizes critique and revision.}
\label{tab:hms-cross-model-clusters}
\end{table}

Applying HMS to Qwen-32B reveals a recurring \textbf{three-function macro-organization} across datasets: \textit{Understanding and Planning} (\textbf{U/P}), \textit{Translating and Drafting} (\textbf{T/D}), and \textit{Refining and Verifying} (\textbf{R/V}); see Appendix~\ref{app:l0-patterns} for their definitions. This organization does not imply an invariant three-cluster solution. The macro-organization remains stable despite dataset-specific variation in the induced clusters; detailed results are provided in Appendix~\ref{app:qwen32b-meta-summaries}.\footnote{These functions are presented sequentially for clarity but may interleave or repeat in actual reasoning traces.}

We further apply HMS to additional LRMs and map their induced clusters to the closest HMS functions in Table~\ref{tab:hms-cross-model-clusters}.
The mapping suggests a broadly shared organization among general-purpose LRMs: Gemma-4-E4B and Llama-8B align closely with U/P--T/D--R/V, while gpt-oss-20B shows partial overlap between drafting and refinement.
DRT-14B differs: its clusters merge initial translation with problem identification and emphasize refinement, revision, evaluation, and critique, suggesting a workflow centered on diagnosis and correction rather than a clean three-stage progression.
Thus, the three-stage HMS hierarchy captures a common structure in general-purpose LRMs, while specialized MT reasoning may reorganize these functions around task-specific revision and critique.

\subsection{Distribution of Reasoning Allocation}

We analyze reasoning allocation on WMT24, whose unified multi-domain setup supports direct cross-domain comparison. We first define two allocation metrics and then examine their distributions across domains and reasoning patterns.

\begin{figure}[tp!]
    \centering
    \begin{subfigure}[b]{\linewidth}
        \centering
        \includegraphics[width=1.0\linewidth]{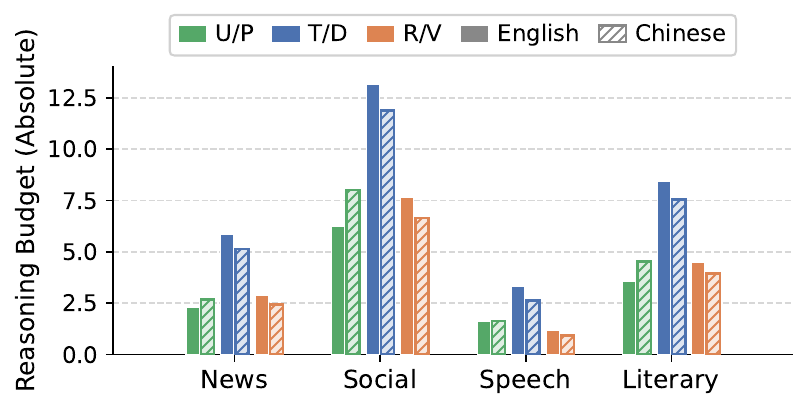}
    \end{subfigure}
    \begin{subfigure}[b]{\linewidth}
        \centering
        \includegraphics[width=1.0\linewidth]{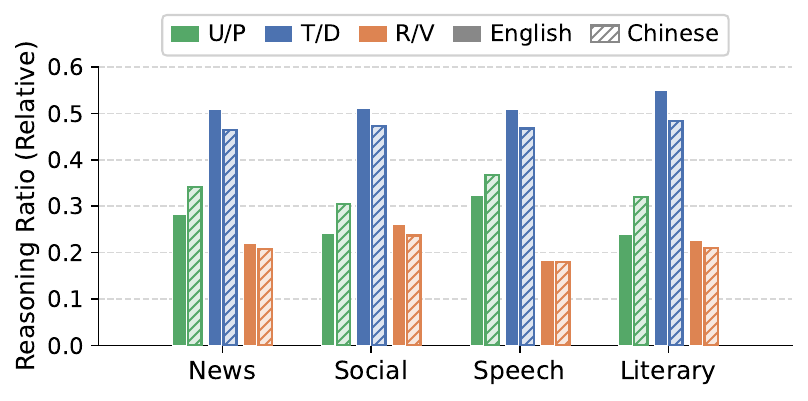}
    \end{subfigure}
    \caption{Level-0 reasoning allocation on WMT24 across four domains. Reasoning ratios remain stable, while reasoning budgets vary modestly by domain.}
    \label{fig:WMT24-L0-Pattern}
\end{figure}

\begin{figure*}[tp!]
    \centering
    \includegraphics[width=1.0\textwidth]{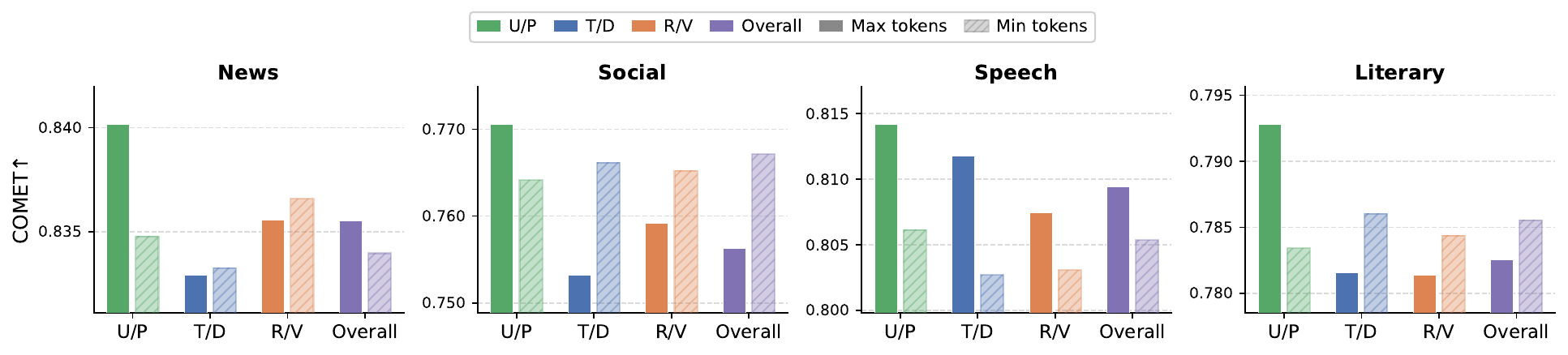}
    \caption{Qwen-32B COMET scores on WMT24 when the reasoning pattern is set to its maximum versus minimum observed length. Longer U/P is the most stable positive signal, while T/D and R/V effects vary more by domain.}
    \label{fig:wmt24_l1_qwen32b_comet}
\end{figure*}

\subsubsection{Reasoning Allocation Metrics}

Using HMS, each reasoning trace is decomposed into atomic reasoning steps, and each step-level summary $z_i$ is assigned hierarchical labels for its Level-0 and Level-1 reasoning patterns. Let $\mathbf{Z}$ denote the set of step-level summaries in a trace, and let $t_i$ be the number of reasoning tokens generated in the corresponding step $s_i$.

We quantify reasoning allocation with two token-based metrics. For a reasoning pattern $p$ (Level-0 or Level-1), the \textbf{Reasoning Budget} is the total number of reasoning tokens assigned to $p$:
\begin{align*}
B(p) = \sum_{z_i \in \mathbf{Z}} t_i \cdot \mathbb{I}\!\left[z_i \in p\right],
\end{align*}
where $\mathbb{I}[\cdot]$ is the indicator function. 
To account for variation in total reasoning length, we define the \textbf{Reasoning Ratio} as the fraction of the overall reasoning budget allocated to $p$:
\begin{align*}
R(p) = \frac{B(p)}{\sum_{z_j \in \mathbf{Z}} t_j}.
\end{align*}
Together, these metrics capture \textit{absolute} and \textit{relative} allocation of reasoning tokens across patterns.

\subsubsection{Domain-General Allocation at Level-0}\label{sec:Level0Allocation}

Figure~\ref{fig:WMT24-L0-Pattern} shows the distribution of Level-0 reasoning patterns for Qwen-32B on WMT24 across four domains, measured by Reasoning Budget and Reasoning Ratio.

\paragraph{Stable Cross-Domain Structure.}
Across domains, reasoning allocation follows a highly consistent structure. Most reasoning is devoted to T/D, while U/P and R/V account for smaller shares. This stability suggests a shared, domain-general allocation of translation reasoning.

\paragraph{Absolute vs. Relative Allocation.}
Reasoning Ratios remain stable across domains, whereas absolute budgets vary: social and literary texts elicit longer traces than news and speech despite having shorter inputs.\footnote{Social and literary inputs average 15.69 and 38.06 words, respectively, compared with 54.05 for news and 72.98 for speech.} This may reflect their greater need for pragmatic, cultural, or stylistic deliberation.

\paragraph{Effect of Reasoning Language.}
Traces produced under English and Chinese instructions share the same high-level U/P-T/D-R/V organization, although Chinese reasoning allocates relatively more budget to U/P and less to T/D and R/V. This suggests that reasoning language redistributes effort without altering the underlying translation process.

\subsubsection{Domain-Specific Allocation at Level-1}

To examine how shared Level-0 stages adapt to domain demands, we analyze Level-1 Reasoning Ratios, which normalize for trace length and reveal within-stage composition. Allocation varies substantially across domains: U/P foregrounds different preparatory cues, T/D shifts between sentence- and phrase/term-level translation, and R/V emphasizes different revision functions. Thus, HMS captures a stable high-level structure alongside domain-sensitive fine-grained allocation. Full results are provided in Appendix~\ref{appendix:domain-specific}.
\subsection{Effectiveness of Reasoning Patterns}

We next examine whether longer function-specific reasoning is associated with better WMT24 translations. Figure~\ref{fig:wmt24_l1_qwen32b_comet} compares Qwen-32B COMET scores at the maximum and minimum observed token allocations for each pattern; additional metrics and model settings appear in Appendix~\ref{app:reasoning-pattern-effectiveness}.

\paragraph{Planning Is Most Reliable (U/P).}
Greater U/P effort shows the clearest and most stable positive association with translation quality across domains, suggesting that additional source understanding and planning help the model identify constraints, anticipate ambiguities, and choose an appropriate strategy.

\paragraph{Drafting Helps Selectively (T/D).}
Longer T/D traces are not uniformly beneficial. The association is strongest for speech, where more elaborate drafting may help handle disfluency, informality, and context-dependent phrasing, but the gains are weaker or less consistent in other domains.

\paragraph{Verification Is Domain-Dependent (R/V).}
Longer R/V traces show the most mixed pattern, aligning with higher scores in some settings but lower scores in others. One interpretation is that extended verification is often triggered by harder examples; another is that excessive revision may lead to over-editing. In either case, verification appears most useful when applied selectively.

Overall, the value of longer reasoning depends strongly on its function: additional U/P is most consistently helpful, T/D helps mainly in reformulation-heavy domains, and R/V is sensitive to domain and example difficulty. These trends persist across additional metrics and model configurations in the appendix.

\section{Conclusion}\label{sec:conclusion}

We systematically analyzed reasoning traces in LRM-based MT, showing that reasoning language, length, and structure jointly shape translation quality and efficiency.
The most effective reasoning language is model-dependent and tends to align with the language a model follows most reliably, while reasoning length has a non-monotonic effect: moderate reasoning can help, but excessive reasoning may degrade quality.
We further introduced \textbf{Hierarchical Meta-Summarization (HMS)}, a data-driven framework for inducing hierarchical reasoning patterns.
HMS reveals a three-stage organization---\textit{Understanding and Planning}, \textit{Translating and Drafting}, and \textit{Refining and Verifying}---shared by general-purpose LRMs at a high level, with domain-specific variation in lower-level allocation and effectiveness.
Together, these findings suggest that uniform reasoning strategies are suboptimal for MT, motivating model-aware, pattern-aware, and domain-sensitive control.

\section*{Limitations}

Conceptually, the identified stages align with prior step-based and agent-based translation frameworks, such as planning-generation-revision pipelines~\cite{briakou2024translating,wang2024drt,he2025r1}.
Unlike these approaches, HMS induces its hierarchy directly from model-generated traces, without predefined stages or agent roles, revealing fine-grained and domain-dependent strategies beyond fixed pipelines.
This provides an empirical basis for adaptive, stage-specific reasoning analysis and control.

This study also leaves several directions for future work.
First, we focus on open-source LRMs with strong multilingual capability and accessible reasoning traces, which are required for HMS but typically unavailable from closed-source systems such as Gemini, Claude, and GPT-5. Extending the analysis to more models, architectures, and low-resource languages would further test the generality of the observed structures.
Second, HMS should be interpreted as a scalable, human-validated analysis framework rather than a model-independent ground-truth taxonomy. Although its abstraction and labeling stages may introduce noise or reflect model biases, our human validation suggests that the resulting summaries and labels are generally reliable. We therefore use HMS as a scalable tool for interpreting reasoning traces, while treating its fine-grained categories as analytical constructs rather than definitive labels.
Finally, because most quality analyses rely on learned MT metrics, the results should be interpreted as metric-aligned quality trends rather than direct human preference judgments.

\section*{Acknowledgments}
We sincerely thank Sweta Agrawal, Elizabeth Nielsen, Dan Deutsch, and Markus Freitag for their insightful discussions, constructive feedback, and valuable suggestions throughout this work. We also thank the members of the Google Translate Research team for sharing their expertise and perspectives. This research was conducted during Yuxiang Liu's internship at Google and benefited from Google's collaborative research environment.

\bibliography{custom}

\appendix

\section{Reasoning Language Analysis}
\subsection{Full Reasoning-Language Results for Qwen-14B}
\label{app:qwen14b-reasoning-language}

This section reports the full Qwen-14B results for reasoning-language adherence (Table~\ref{tab:qwen14b-instruction-adherence}) and translation quality (Table~\ref{tab:qwen14b-quality-by-inst-lang}) across dataset-language pairs. These results complement the main analysis, showing that Chinese reasoning instructions achieve higher overall adherence and slightly higher mean translation quality for Qwen models, although English remains stronger for some English-to-European language pairs.

\begin{table*}[tp!]
    \centering
    \small
    \setlength{\tabcolsep}{5.5pt}
    \begin{tabular}{lcccccccccc}
        \toprule
        \textbf{Inst.$\rightarrow$Reason}
        & \multicolumn{5}{c}{\textbf{WMT}}
        & \multicolumn{3}{c}{\textbf{CMT}}
        & \textbf{DRT}
        & \textbf{Mean} \\
        \cmidrule(lr){2-6} \cmidrule(lr){7-9} \cmidrule(lr){10-10}
        & \textbf{en-de} & \textbf{en-es} & \textbf{en-ja} & \textbf{en-ru} & \textbf{en-zh}
        & \textbf{en-es} & \textbf{en-fr} & \textbf{en-zh}
        & \textbf{en-zh}
        & \\
        \midrule
        en$\rightarrow$en (\%)
        & 87.7 & \textbf{97.3} & 43.5 & 72.8 & 71.6
        & \textbf{95.9} & \textbf{96.7} & 64.0
        & 66.8 & 77.4 \\
        zh$\rightarrow$zh (\%)
        & \textbf{93.2} & 91.2 & \textbf{93.9} & \textbf{82.4} & \textbf{97.6}
        & 90.0 & 90.4 & \textbf{99.2}
        & \textbf{98.7} & \textbf{92.9} \\
        \bottomrule
    \end{tabular}

    \caption{Reasoning-Language Instruction-Adherence (RLIA) rates for Qwen-14B. Chinese instructions yield higher overall adherence than English, though English is stronger on several English-to-European language pairs.}
    \label{tab:qwen14b-instruction-adherence}
\end{table*}

\begin{table*}[tp!]
    \centering
    \small
    \setlength{\tabcolsep}{4.2pt}
    \begin{tabular}{llcccccccccc}
        \toprule
        \textbf{Metric} & \textbf{Inst.}
        & \multicolumn{5}{c}{\textbf{WMT}}
        & \multicolumn{3}{c}{\textbf{CMT}}
        & \textbf{DRT}
        & \textbf{Mean} \\
        \cmidrule(lr){3-7} \cmidrule(lr){8-10} \cmidrule(lr){11-11}
        &
        & \textbf{en-de} & \textbf{en-es} & \textbf{en-ja} & \textbf{en-ru} & \textbf{en-zh}
        & \textbf{en-es} & \textbf{en-fr} & \textbf{en-zh}
        & \textbf{en-zh}
        & \\
        \midrule
        \multirow{2}{*}{COMET$\uparrow$}
        & en & \textbf{0.7520} & 0.7872 & 0.8051 & \textbf{0.7259} & 0.8227 & \textbf{0.8205} & \textbf{0.7713} & 0.8267 & 0.7600 & 0.7857 \\
        & zh & 0.7450 & \textbf{0.7940} & \textbf{0.8086} & 0.7192 & \textbf{0.8285} & 0.8176 & 0.7684 & \textbf{0.8329} & \textbf{0.7669} & \textbf{0.7868} \\
        \midrule
        \multirow{2}{*}{Norm.MX$\uparrow$}
        & en & 0.8258 & 0.8071 & 0.7334 & 0.6925 & 0.8644 & 0.8572 & 0.8168 & 0.8420 & 0.8163 & 0.8062 \\
        & zh & \textbf{0.8346} & \textbf{0.8282} & \textbf{0.7555} & \textbf{0.7076} & \textbf{0.8749} & \textbf{0.8609} & \textbf{0.8175} & \textbf{0.8565} & \textbf{0.8272} & \textbf{0.8181} \\
        \midrule
        \multirow{2}{*}{Norm.MX-QE$\uparrow$}
        & en & 0.8438 & 0.8138 & 0.7508 & 0.7204 & 0.8730 & 0.8749 & 0.8648 & 0.8579 & 0.8164 & 0.8240 \\
        & zh & \textbf{0.8543} & \textbf{0.8420} & \textbf{0.7745} & \textbf{0.7380} & \textbf{0.8838} & \textbf{0.8793} & \textbf{0.8659} & \textbf{0.8729} & \textbf{0.8282} & \textbf{0.8376} \\
        \bottomrule
    \end{tabular}

    \caption{Translation quality for Qwen-14B by instructed reasoning language. Chinese reasoning yields higher mean scores across all three metrics, with especially consistent gains on Norm.MX and Norm.MX-QE.}
    \label{tab:qwen14b-quality-by-inst-lang}
\end{table*}

\subsection{Effect of Reasoning-Language Control}
\label{app:reasoning-language-control}

Our experiments in Section~\ref{sec:language} showed that the choice of reasoning language matters, but did not directly establish whether explicitly controlling the reasoning language is preferable to leaving it unconstrained. We therefore evaluate a \textit{no-language-control baseline} on WMT24 en-zh for Qwen-14B, Qwen-32B, and Llama-8B. This baseline uses the same decoding configuration and translation prompt but omits the reasoning-language instruction.

\begin{table*}[t]
\centering
\begin{tabular}{llccccc}
\toprule
\textbf{Model} & \textbf{Inst.} & \textbf{zh\%} & \textbf{en\%} &
\textbf{COMET $\uparrow$} & \textbf{Norm.MX $\uparrow$} &
\textbf{Norm.MX-QE $\uparrow$} \\
\midrule

\multirow{3}{*}{Qwen-14B}
& English & 28.4\% & 71.6\% & 0.8227 & 0.8644 & 0.8730 \\
& Chinese & 97.6\% & 2.4\% & \textbf{0.8285} & \textbf{0.8749} & \textbf{0.8838} \\
& None    & 38.0\% & 62.0\% & 0.8235 & 0.8659 & 0.8745 \\
\midrule

\multirow{3}{*}{Qwen-32B}
& English & 35.8\% & 64.2\% & 0.8297 & 0.8825 & 0.8884 \\
& Chinese & 99.2\% & 0.8\% & \textbf{0.8363} & \textbf{0.8894} & \textbf{0.8960} \\
& None    & 44.0\% & 56.0\% & 0.8306 & 0.8834 & 0.8894 \\
\midrule

\multirow{3}{*}{Llama-8B}
& English & 22.2\% & 77.8\% & 0.7939 & 0.8374 & 0.8510 \\
& Chinese & 99.9\% & 0.1\% & \textbf{0.8035} & \textbf{0.8468} & \textbf{0.8612} \\
& None    & 30.0\% & 70.0\% & 0.7949 & 0.8383 & 0.8520 \\
\bottomrule
\end{tabular}
\caption{Reasoning-language distributions and translation quality under English, Chinese, and no-language-control conditions on WMT24 en-zh.}
\label{tab:model_instruction_results}
\end{table*}

As shown in Table~\ref{tab:model_instruction_results}, without language control, all three models reason predominantly in English (56.0-70.0\%), possibly because the translation instruction itself is written in English. Explicitly requiring Chinese shifts the reasoning almost entirely to Chinese (97.6-99.9\%) and yields higher scores than the unconstrained baseline across all three metrics for all three models. In contrast, explicitly requiring English produces slightly lower scores than unconstrained reasoning in every model-metric combination.

These results show that reasoning-language control can be beneficial, but the benefit does not arise merely from imposing a language; it depends on selecting an appropriate language for the model. Moreover, a model's unconstrained reasoning-language preference does not necessarily maximize translation quality.
\section{Reasoning Length Analysis}

\subsection{Reasoning Trace Length Statistics}
\label{app:reasoning-length-stats}

This section reports detailed reasoning-trace length statistics used in the reasoning-length analysis. Table~\ref{tab:reasoning-length-stats} summarizes ranked trace lengths across WMT, CMT, and DRT for the Qwen models, while Table~\ref{tab:wmt24pp-reasoning-length-stats} reports the corresponding WMT (en-zh) statistics across all evaluated models. For each input, we sort 16 sampled reasoning traces by length, allowing us to compare naturally short and long traces while controlling for source-sentence difficulty.

\begin{table*}[tp!]
\centering
\scriptsize
\setlength{\tabcolsep}{2.2pt}
\renewcommand{\arraystretch}{1.08}
\resizebox{\textwidth}{!}{
\begin{tabular}{@{}lll*{16}{r}r@{}}
\toprule
\multirow{2}{*}{Dataset}
& \multirow{2}{*}{Model}
& \multirow{2}{*}{Lang.}
& \multicolumn{16}{c}{Ranked reasoning trace length}
& \multirow{2}{*}{\(r_{16}/r_1\)} \\
\cmidrule(lr){4-19}
& &
& \(r_1\) & \(r_2\) & \(r_3\) & \(r_4\) & \(r_5\) & \(r_6\) & \(r_7\) & \(r_8\)
& \(r_9\) & \(r_{10}\) & \(r_{11}\) & \(r_{12}\) & \(r_{13}\) & \(r_{14}\) & \(r_{15}\) & \(r_{16}\)
& \\
\midrule

\multirow{4}{*}{WMT}
& \multirow{2}{*}{Qwen-14B}
& en & 236 & 274 & 300 & 322 & 342 & 361 & 380 & 399 & 419 & 443 & 467 & 495 & 530 & 577 & 648 & 825  & \(3.49\times\) \\
& & zh & 339 & 404 & 444 & 476 & 502 & 527 & 550 & 573 & 596 & 619 & 643 & 671 & 705 & 744 & 800 & 934  & \(2.76\times\) \\
\cmidrule(lr){2-20}
& \multirow{2}{*}{Qwen-32B}
& en & 276 & 325 & 359 & 389 & 416 & 442 & 469 & 496 & 525 & 556 & 592 & 635 & 685 & 752 & 856 & 1092 & \(3.96\times\) \\
& & zh & 406 & 479 & 524 & 559 & 589 & 618 & 644 & 670 & 696 & 724 & 753 & 785 & 826 & 874 & 946 & 1147 & \(2.83\times\) \\

\midrule

\multirow{4}{*}{CMT}
& \multirow{2}{*}{Qwen-14B}
& en & 240 & 277 & 302 & 323 & 341 & 360 & 377 & 395 & 414 & 435 & 457 & 483 & 515 & 554 & 613 & 739  & \(3.07\times\) \\
& & zh & 348 & 407 & 442 & 471 & 495 & 517 & 538 & 559 & 580 & 603 & 626 & 653 & 684 & 722 & 775 & 881  & \(2.53\times\) \\
\cmidrule(lr){2-20}
& \multirow{2}{*}{Qwen-32B}
& en & 277 & 324 & 357 & 382 & 405 & 427 & 450 & 472 & 497 & 522 & 551 & 584 & 622 & 670 & 746 & 894  & \(3.22\times\) \\
& & zh & 391 & 454 & 494 & 525 & 552 & 578 & 602 & 627 & 651 & 677 & 705 & 736 & 772 & 818 & 883 & 1035 & \(2.65\times\) \\

\midrule

\multirow{4}{*}{DRT}
& \multirow{2}{*}{Qwen-14B}
& en & 229 & 262 & 286 & 305 & 323 & 339 & 355 & 372 & 390 & 408 & 428 & 451 & 479 & 514 & 562 & 656  & \(2.86\times\) \\
& & zh & 386 & 447 & 484 & 513 & 537 & 561 & 584 & 605 & 626 & 649 & 672 & 698 & 728 & 763 & 811 & 905  & \(2.35\times\) \\
\cmidrule(lr){2-20}
& \multirow{2}{*}{Qwen-32B}
& en & 241 & 282 & 311 & 336 & 358 & 379 & 400 & 420 & 442 & 465 & 491 & 519 & 555 & 599 & 663 & 804  & \(3.34\times\) \\
& & zh & 417 & 485 & 529 & 564 & 594 & 622 & 647 & 673 & 699 & 725 & 754 & 786 & 823 & 867 & 932 & 1075 & \(2.58\times\) \\

\bottomrule
\end{tabular}
}
\caption{Reasoning trace length by rank across datasets. For each input, 16 traces are sorted from shortest (\(r_1\)) to longest (\(r_{16}\)); cells report the mean length at each rank.}
\label{tab:reasoning-length-stats}
\end{table*}

\begin{table*}[tp!]
\centering
\scriptsize
\setlength{\tabcolsep}{2.4pt}
\renewcommand{\arraystretch}{1.08}
\resizebox{\textwidth}{!}{
\begin{tabular}{@{}ll*{16}{r}r@{}}
\toprule
\multirow{2}{*}{Model}
& \multirow{2}{*}{Lang.}
& \multicolumn{16}{c}{Ranked reasoning trace length}
& \multirow{2}{*}{\(r_{16}/r_1\)} \\
\cmidrule(lr){3-18}
& 
& \(r_1\) & \(r_2\) & \(r_3\) & \(r_4\) & \(r_5\) & \(r_6\) & \(r_7\) & \(r_8\)
& \(r_9\) & \(r_{10}\) & \(r_{11}\) & \(r_{12}\) & \(r_{13}\) & \(r_{14}\) & \(r_{15}\) & \(r_{16}\)
& \\
\midrule

\multirow{2}{*}{Qwen-14B}
& en & 195 & 226 & 249 & 268 & 284 & 301 & 319 & 335 & 351 & 377 & 396 & 416 & 442 & 473 & 516 & 599 & \textbf{\(3.07\times\)} \\
& zh & 324 & 380 & 418 & 448 & 473 & 497 & 518 & 538 & 559 & 580 & 603 & 632 & 665 & 698 & 740 & 817 & \(2.52\times\) \\
\cmidrule(lr){1-19}

\multirow{2}{*}{Qwen-32B}
& en & 212 & 253 & 281 & 305 & 327 & 348 & 369 & 392 & 416 & 439 & 466 & 496 & 531 & 575 & 655 & 796 & \textbf{\(3.76\times\)} \\
& zh & 369 & 432 & 478 & 513 & 543 & 572 & 598 & 625 & 649 & 675 & 702 & 730 & 765 & 807 & 866 & 996 & \(2.70\times\) \\
\cmidrule(lr){1-19}

gpt-oss-20B
& en & 158 & 204 & 240 & 269 & 296 & 318 & 340 & 364 & 385 & 408 & 433 & 460 & 490 & 530 & 587 & 704 & \(4.45\times\) \\
\cmidrule(lr){1-19}

\multirow{2}{*}{Llama-8B}
& en & 237 & 276 & 304 & 327 & 347 & 368 & 389 & 410 & 428 & 450 & 473 & 498 & 528 & 563 & 610 & 705 & \textbf{\(2.98\times\)} \\
& zh & 287 & 346 & 384 & 412 & 436 & 458 & 478 & 499 & 520 & 541 & 564 & 591 & 620 & 655 & 706 & 801 & \(2.79\times\) \\
\cmidrule(lr){1-19}

Gemma-4-E4B
& en & 460 & 544 & 609 & 660 & 695 & 724 & 752 & 778 & 801 & 822 & 842 & 866 & 891 & 922 & 959 & 1023 & \(2.23\times\) \\

\bottomrule
\end{tabular}
}
\caption{Reasoning trace length by rank on WMT (en-zh). For each input, 16 traces are sorted from shortest (\(r_1\)) to longest (\(r_{16}\)); cells report the mean length at each rank.}
\label{tab:wmt24pp-reasoning-length-stats}
\end{table*}

\subsection{Quality by Reasoning-Length Rank}
\label{app:reasoning-length-quality-curves}

\begin{figure*}[tp!]
    \centering
    \includegraphics[width=1.0\textwidth]{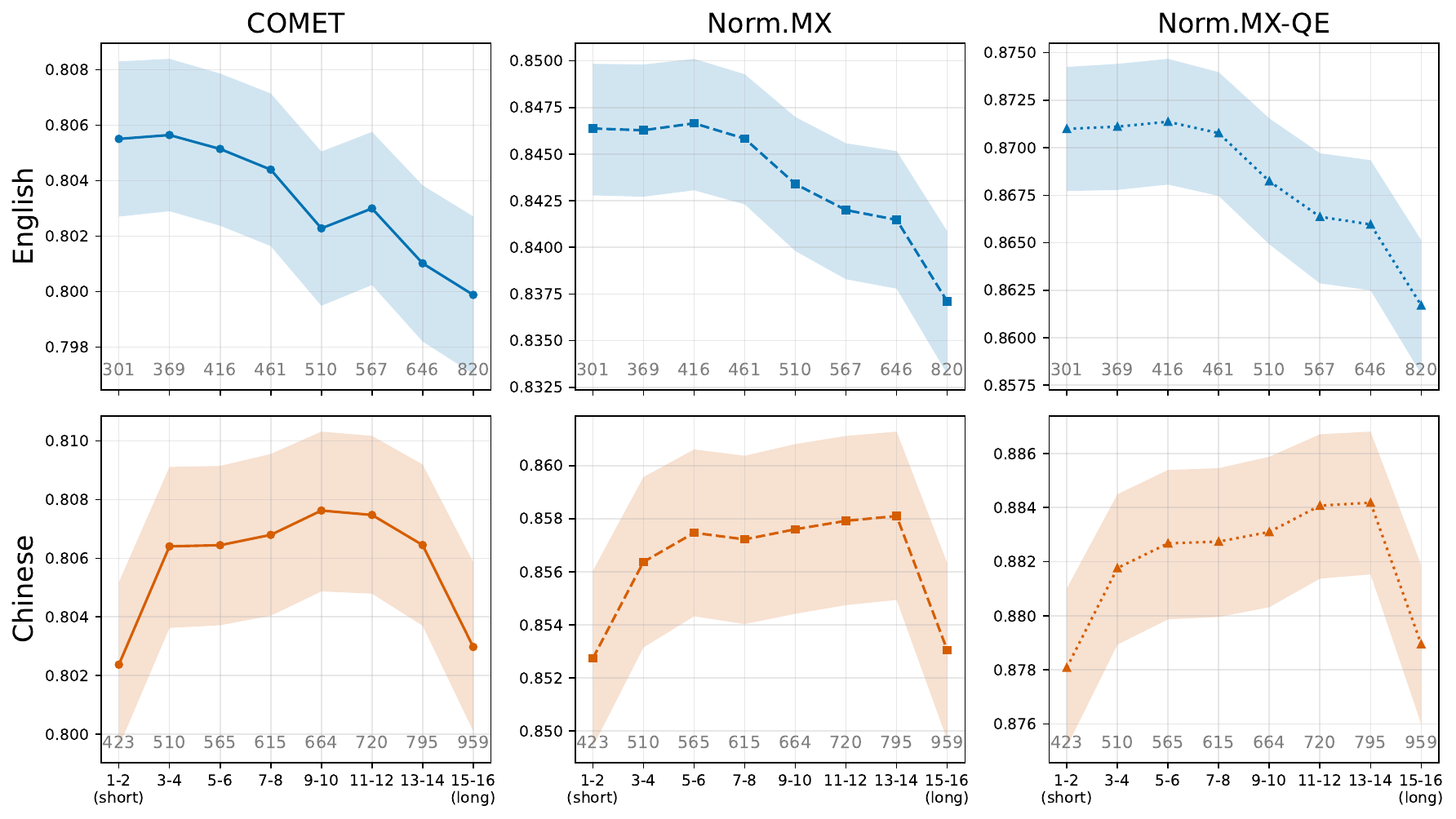}
    \caption{Translation quality by ranked reasoning trace length for Qwen-32B on CMT.}
    \label{fig:cmt-qwen32B-reasoning-length}
\end{figure*}

\begin{figure*}[tp!]
    \centering
    \includegraphics[width=1.0\textwidth]{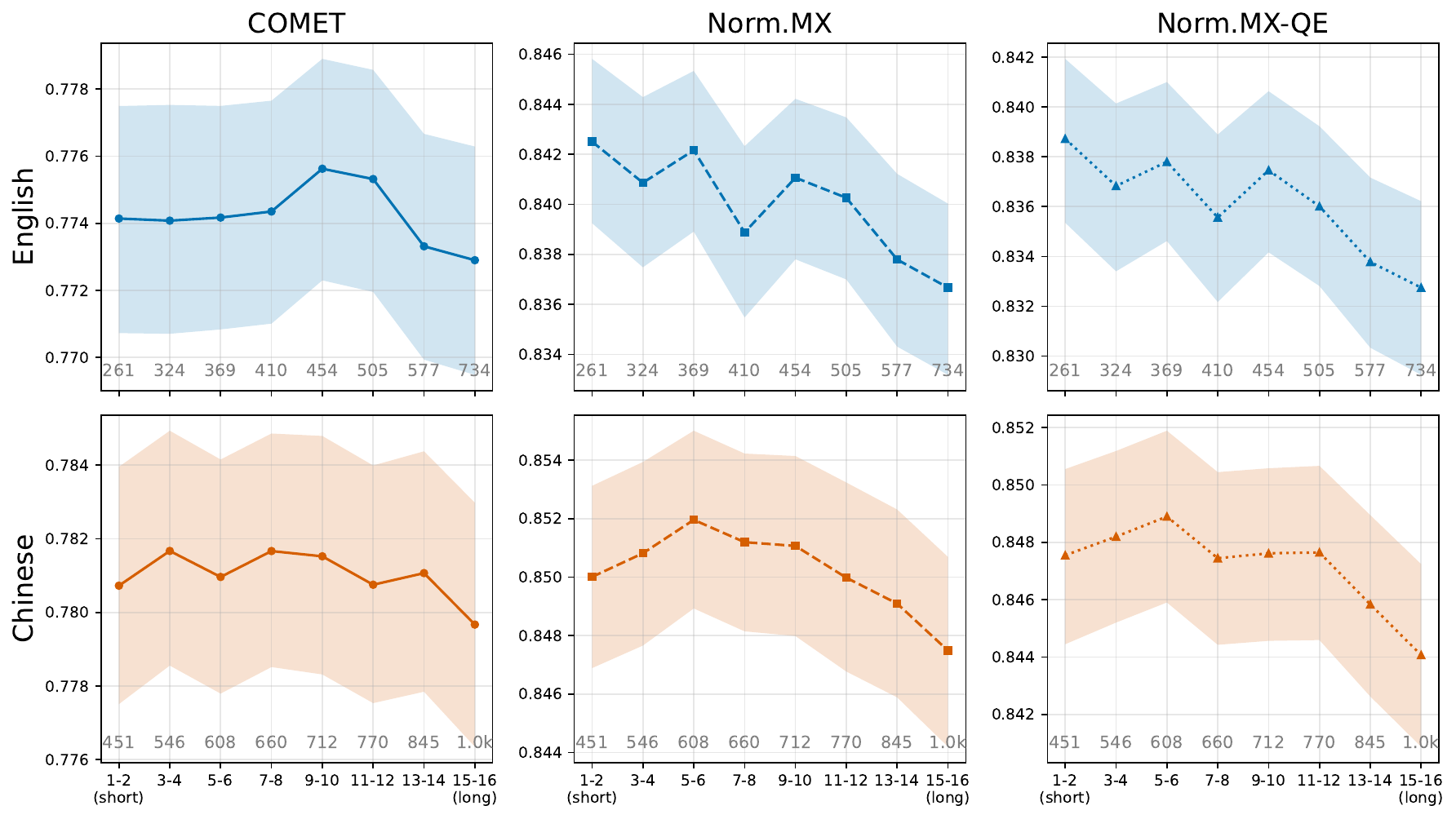}
    \caption{Translation quality by ranked reasoning trace length for Qwen-32B on DRT.}
    \label{fig:drt-qwen32B-reasoning-length}
\end{figure*}

\begin{figure*}[tp!]
    \centering
    \includegraphics[width=1.0\textwidth]{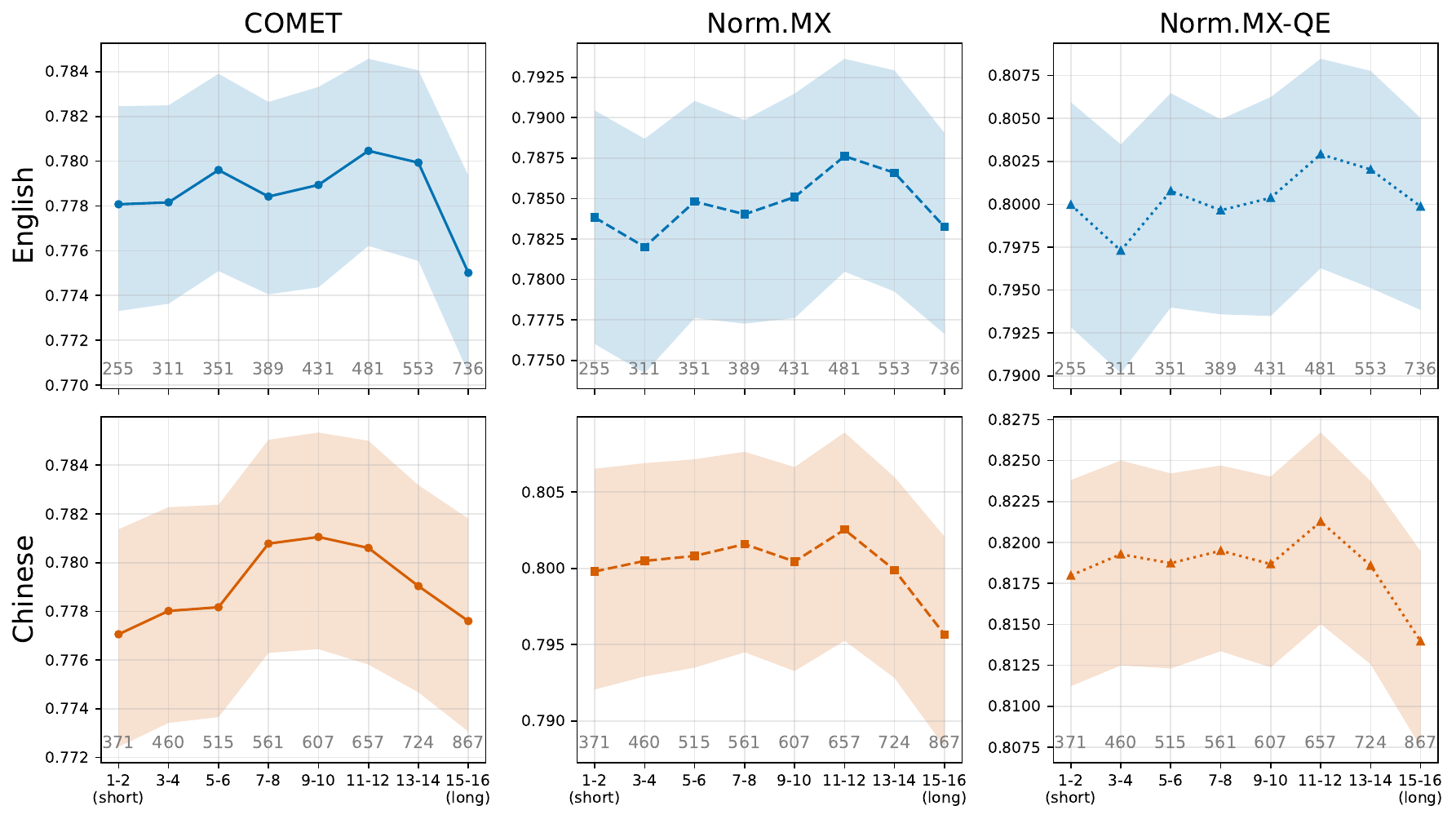}
    \caption{Translation quality by ranked reasoning trace length for Qwen-14B on WMT.}
    \label{fig:wmt24-qwen14B-reasoning-length}
\end{figure*}

\begin{figure*}[tp!]
    \centering
    \includegraphics[width=1.0\textwidth]{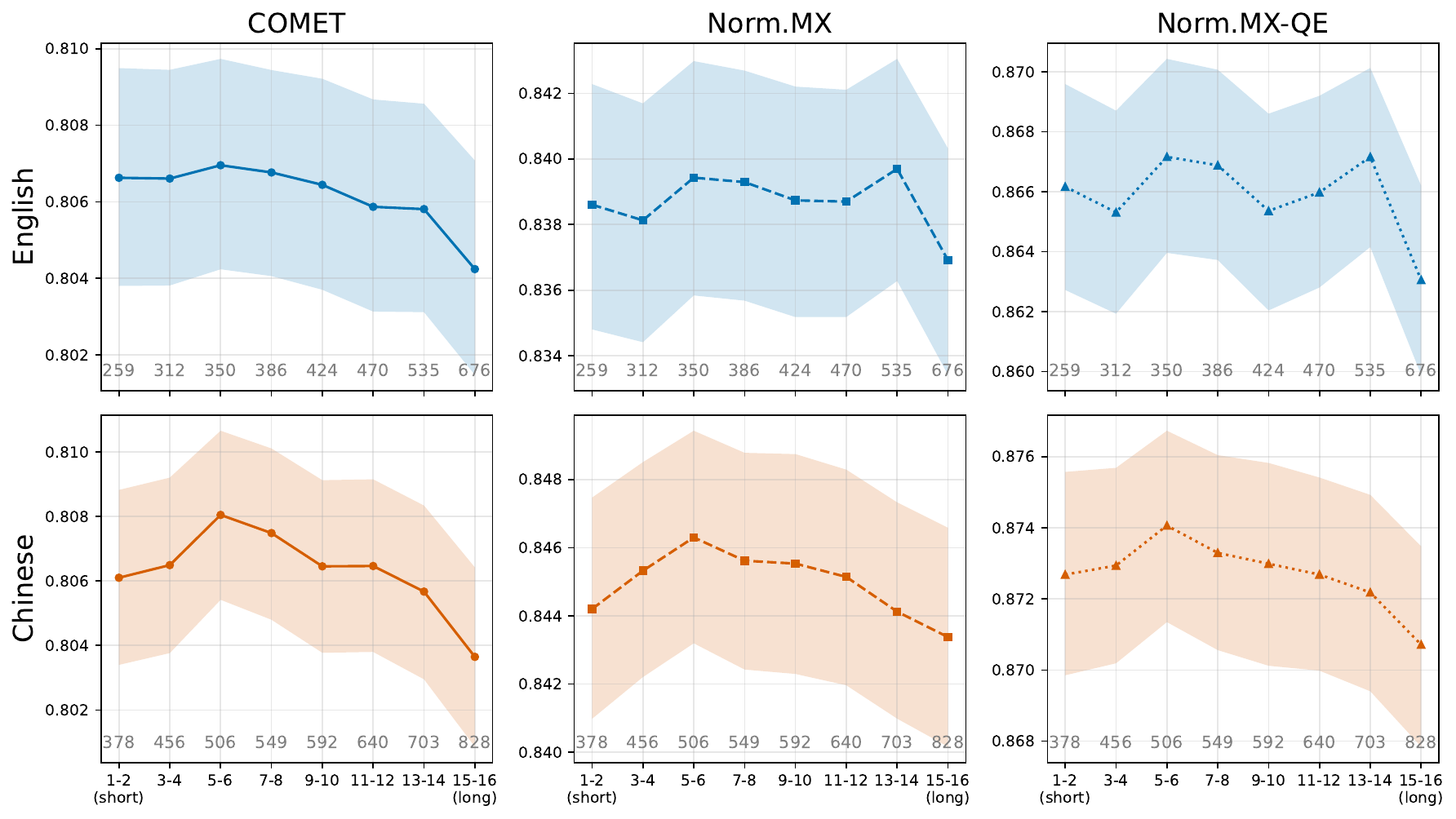}
    \caption{Translation quality by ranked reasoning trace length for Qwen-14B on CMT.}
    \label{fig:cmt-qwen14B-reasoning-length}
\end{figure*}

\begin{figure*}[tp!]
    \centering
    \includegraphics[width=1.0\textwidth]{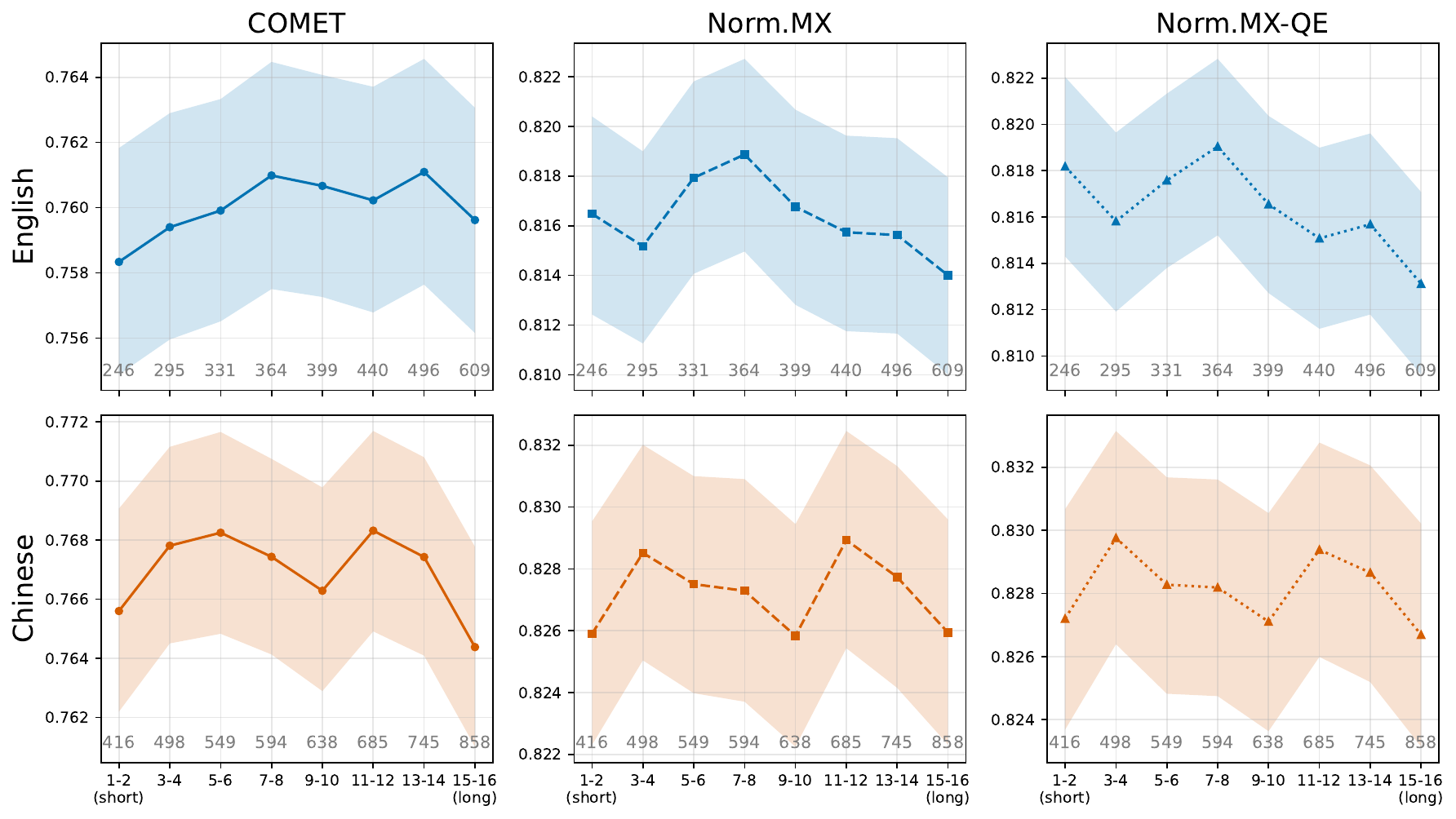}
    \caption{Translation quality by ranked reasoning trace length for Qwen-14B on DRT.}
    \label{fig:drt-qwen14B-reasoning-length}
\end{figure*}

\begin{figure*}[tp!]
    \centering
    \includegraphics[width=1.0\textwidth]{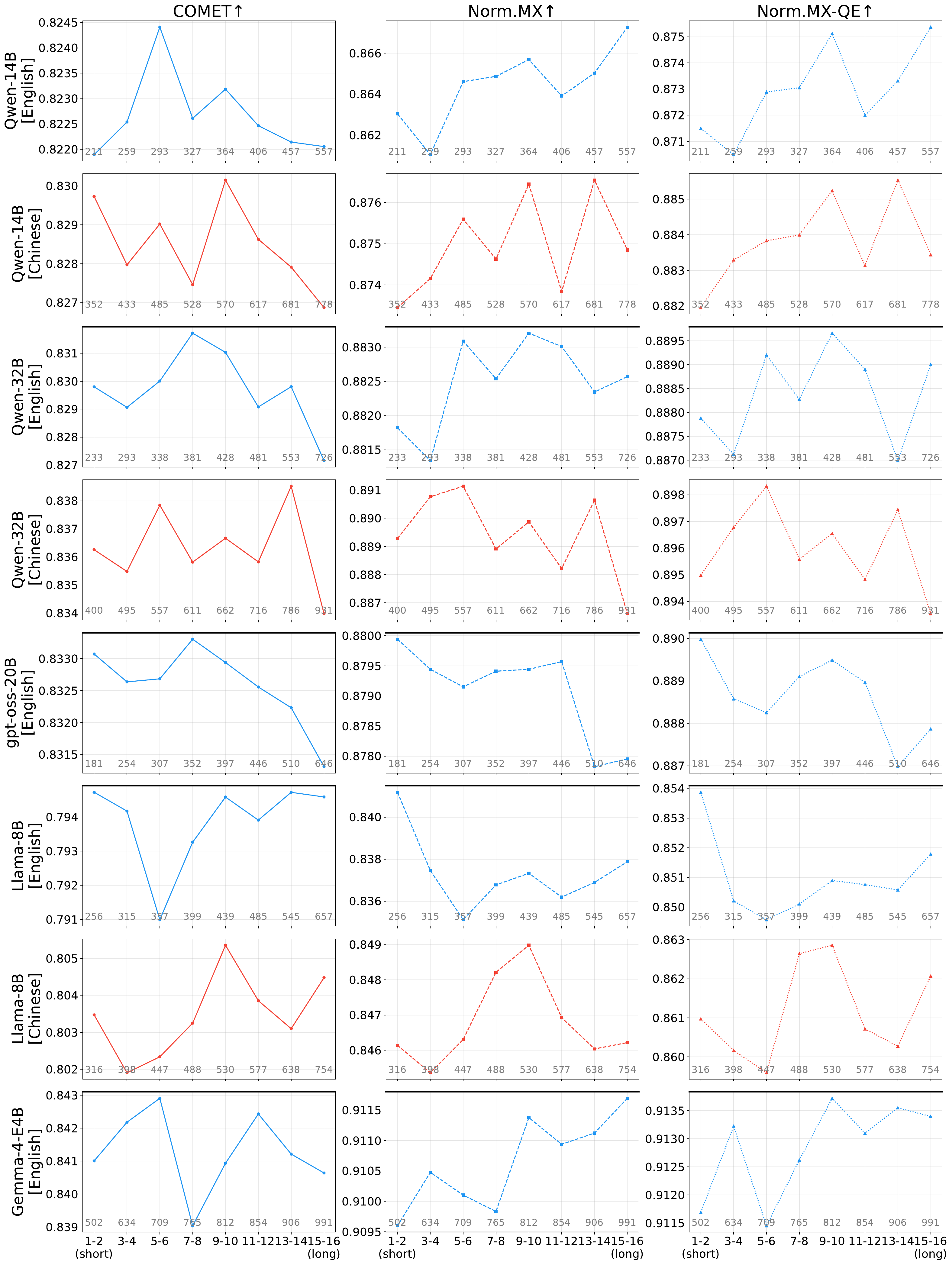}
    \caption{Translation quality by ranked reasoning trace length on WMT (en-zh) across evaluated models. Adjacent length ranks are grouped into bins from shortest to longest traces.}
    \label{fig:wmt24-en_zh-reasoning-length}
\end{figure*}

This section provides the full set of ranked-length quality curves complementing the main discussion in Section~\ref{sec:ReasoningEfficiency}.
Figures~\ref{fig:cmt-qwen32B-reasoning-length}, \ref{fig:drt-qwen32B-reasoning-length}, \ref{fig:wmt24-qwen14B-reasoning-length}, \ref{fig:cmt-qwen14B-reasoning-length}, \ref{fig:drt-qwen14B-reasoning-length}, and~\ref{fig:wmt24-en_zh-reasoning-length} group the 16 sampled traces for each source sentence by length rank, from shortest to longest, and report the translation quality associated with each rank group.
Across datasets and models, these curves show that quality does not consistently improve with longer reasoning: intermediate-length traces can be competitive or beneficial, whereas the longest traces often provide limited gains and may degrade performance.

\subsection{Reasoning-Effort Control}
\label{app:reasoning-length-control}

To complement the observational analysis in Section~\ref{sec:ReasoningEfficiency}, we evaluate whether explicitly increasing reasoning effort improves translation quality.
Unlike the ranked-length analysis, which compares \textit{naturally} short and long traces sampled from the same model, this experiment directly \textit{controls} the model's reasoning budget.
We run gpt-oss-20B on WMT (en-zh) under low, medium, and high reasoning-effort settings, reporting both translation quality and the resulting average trace length.

Table~\ref{tab:gpt-oss-effort-wmt24pp} shows that higher effort substantially increases reasoning length, from 93.3 words under low effort to 173.1 words under high effort.
The quality gains, however, are modest: Norm.MX and Norm.MX-QE improve slightly with higher effort, whereas COMET peaks at medium effort.
These results suggest that controlled additional reasoning can be beneficial, although the gains are small and metric-dependent.
Together with the ranked-length results, this finding supports controlling reasoning length adaptively rather than increasing it indiscriminately.

\begin{table}[tp!]
\centering
\small
\setlength{\tabcolsep}{6pt}
\renewcommand{\arraystretch}{1.08}
\begin{tabular}{@{}lrrrr@{}}
\toprule
Effort & COMET & Norm.MX & Norm.MX-QE & \#words \\
\midrule
Low    & 0.8665 & 0.8428 & 0.8501 &  93.3 \\
Med & \textbf{0.8696} & 0.8455 & 0.8530 & 143.6 \\
High   & 0.8689 & \textbf{0.8469} & \textbf{0.8547} & 173.1 \\
\bottomrule
\end{tabular}
\caption{Controlled reasoning effort for gpt-oss-20B on WMT (en-zh). Higher effort increases trace length and slightly improves Norm.MX and Norm.MX-QE, while COMET peaks at medium effort.}
\label{tab:gpt-oss-effort-wmt24pp}
\end{table}

\section{API and Token Costs for HMS}
\label{app:hms-cost-breakdown}

Table~\ref{tab:hms-costs} reports the approximate per-trace API and token costs for the HMS processing pipeline. We separate costs into three stages: step segmentation and abstraction, meta-summarization, and step labeling. Because meta-summarization is amortized across traces, its per-trace contribution is negligible. The dominant cost comes from step labeling, which requires repeated calls over the segmented trace representation.

\begin{table*}[tp!]
\centering
\resizebox{\textwidth}{!}{%
\begin{tabular}{lrrr}
\toprule
\textbf{HMS stage} & \textbf{API calls per trace} & \textbf{Input tokens per trace} & \textbf{Output tokens per trace} \\
\midrule
Step segmentation + abstraction & 1 & $\sim$900--1{,}000 & $\sim$570--690 \\
Meta-summarization & amortized, negligible & $<10$ & $<10$ \\
Step labeling & $\sim$7 & $\sim$5{,}300--6{,}500 & $\sim$15--30 \\
\midrule
\textbf{Total, step-wise labeling} & \textbf{$\sim$8 calls} & \textbf{$\sim$6.2k--7.5k} & \textbf{$\sim$0.6k--0.7k} \\
\bottomrule
\end{tabular}
}
\caption{Approximate API call and token costs per trace for HMS step-wise labeling.}
\label{tab:hms-costs}
\end{table*}
\section{Validation of HMS Outputs}
\label{app:hms-validation}

HMS relies on LLM-based abstraction and labeling to summarize reasoning steps and assign hierarchical labels. To assess the reliability of these outputs, we conducted validation along three dimensions: summary faithfulness, label correctness, and stability across random seeds. The results indicate that HMS outputs are generally faithful and interpretable, while also highlighting the need to treat HMS as an analysis framework rather than a model-independent ground-truth taxonomy.

\begin{table}[htbp]
\centering
\begin{tabular}{lcc}
\toprule
\textbf{Evaluation} & \textbf{Yes rate} & \textbf{Fleiss' $\kappa$} \\
\midrule
Summary faithfulness      & 92\% & 0.71 \\
Level-0 label correctness & 88\% & 0.68 \\
Level-1 label correctness & 79\% & 0.61 \\
R/V corrective behavior   & 67\% & 0.65 \\
R/V checking behavior     & 24\% & 0.60 \\
Overall R/V validity      & 91\% & 0.74 \\
\bottomrule
\end{tabular}
\caption{Human evaluation of HMS summary and label quality with inter-annotator agreement.}
\label{tab:hms-human-validation}
\end{table}

\subsection{Human Validation of Meta Summaries}
\label{app:hms-human-eval}

We randomly sampled 200 reasoning steps across datasets, reasoning languages, models, and HMS categories. For each instance, three annotators\footnote{We hired three graduate students with MT, multilingual NLP, or linguistic annotation experience, compensated them above the local minimum wage, and provided written guidelines and examples.} were shown the source sentence, the original reasoning step, the HMS-generated summary, and the assigned Level-0 and Level-1 HMS labels. Annotators evaluated each instance along two dimensions:

\noindent \textbf{(1) Summary faithfulness:} whether the HMS-generated summary faithfully captures the content and function of the original reasoning step. Summaries were rated as accurate/complete or incorrect/misleading.

\noindent \textbf{(2) Label correctness:} whether the assigned Level-0 and Level-1 HMS labels correctly describe the function of the reasoning step. Labels were rated as correct or incorrect.

Annotations were aggregated by majority vote. As shown in Table~\ref{tab:hms-human-validation}, 92\% of summaries were judged accurate and complete; and for label correctness, 88\% of Level-0 labels and 79\% of Level-1 labels were judged correct. The lower correctness rate for Level-1 labels reflects the greater ambiguity of fine-grained reasoning functions, whose category boundaries are less distinct than those of the coarse-grained Level-0 layer.


\subsection{Human Validation of R/V Pattern}
\label{app:hms-rv-validation}

We further conducted a targeted validation of steps labeled as \textit{Refining and Verifying} (R/V), as revision and verification behavior has received broad attention in analyses of reasoning and self-correction. This analysis examines whether HMS-labeled R/V steps correspond to genuine revision or verification behavior in the translation process.

For each sampled R/V step, annotators inspected the source text, the reasoning step, and the final translation. They judged whether the step performed refinement or verification, whether it identified a concrete translation issue, and whether that issue was reflected in the final output. Each R/V-labeled step was assigned to one of three categories:

\begin{itemize}
    \item \textbf{Corrective}: the step identified or corrected a concrete translation issue that was reflected in the final translation.
    \item \textbf{Checking}: the step performed a valid verification or checking function but did not lead to an observable change in the final translation.
    \item \textbf{Incorrect}: the step did not perform valid refinement or verification, or it introduced an incorrect change.
\end{itemize}

We define ``overall R/V validity'' as the union of corrective and checking behaviors: a step is valid if it performs either function. As shown in Table~\ref{tab:hms-human-validation}, 67\% of R/V-labeled steps identified or corrected a concrete translation issue reflected in the final output, while 24\% performed verification without an observable final change. Together, these categories account for 91\% of R/V-labeled steps, indicating that most genuinely performed refinement or verification. This finding suggests that HMS-identified R/V steps typically capture meaningful revision or checking behavior rather than superficial restatement or irrelevant reasoning.

\subsection{Stability of HMS Patterns}
\label{app:hms-stability}

To evaluate the stability of HMS patterns, we reran the HMS pipeline with different random seeds and compared the resulting high-level structures and category allocation trends. Across runs, the same three Level-0 categories were consistently recovered: \textit{Understanding and Planning}, \textit{Translating and Drafting}, and \textit{Refining and Verifying}. The main domain-level allocation trends also remained stable.

These results suggest that the high-level HMS structure and aggregate reasoning patterns are not artifacts of a single random initialization. However, some variation may remain at finer levels of the hierarchy, particularly for Level-1 labels whose functional distinctions are more ambiguous.

\clearpage
\onecolumn
\section{Prompts}
\label{appendix:prompts}

\tcbset{
  promptbox/.style={
    findingbox,
    breakable,
    width=\textwidth,
    before skip=4pt,
    after skip=8pt,
    left=7pt,
    right=7pt,
    top=4pt,
    bottom=4pt,
    boxsep=2pt,
    fontupper=\footnotesize,
    fonttitle=\small\bfseries
  }
}

\subsection{Reasoning Trace Abstraction}
\label{prompt:segment&summarize}

This prompt segments each raw trace into reasoning steps and functional summaries.

\phantomsection\label{prompt:box-segment-summarize}
\begingroup
\begin{tcolorbox}[promptbox, title=Prompt for Reasoning Trace Abstraction]
ROLE:\\
You are an expert AI assistant specializing in the semantic analysis and structuring of language model reasoning processes. Your goal is to make a complex, raw thinking trace understandable by segmenting it into logical, high-level steps.\\

TASK:\\
You are given a complete thinking trace as a single block of text. Your task is to semantically segment this text into a structured set of coherent reasoning steps. This involves two main actions:\\
1. Segmenting: Read through the entire text and identify the logical breakpoints. Group consecutive sentences or paragraphs that are semantically connected and work towards a single sub-goal.\\
2. Summarizing: Assign a concise, functional title to each new reasoning step that describes its purpose.\\

INPUT FORMAT:\\
A single multi-line string containing the complete, raw thinking trace.\\

OUTPUT FORMAT:\\
Your output MUST be a single, valid JSON object. The keys should be "s0", "s1", "s2", etc., representing the sequential order of the steps. Each value must be an object with two keys: "title" and "content", with an appropriate "title" summarizing its logical purpose and "content" containing the merged, cleaned reasoning text. Please ensure the output is structured, coherent, and well-suited for subsequent semantic analysis or graph-based modeling of the reasoning process. Do not include any text or markdown formatting before or after the JSON object, as an example below:\\
\{
 "s0": \{
     "title": "...",
     "content": "..."
\},
 "s1": \{
     "title": "...",
     "content": "..."
\},
 ...
\}\\

CRITERIA FOR SEGMENTATION:\\
- Semantic Cohesion: Group related sentences or paragraphs that collectively express a single coherent sub-task, logical inference, or a closely related set of thoughts. For example, all text that describes the initial problem setup should be one group. All text that executes a specific line of reasoning should be another.\\
- Self-Contained Steps: Ensure each resulting step has enough context to be understood independently, but avoid merging unrelated steps.\\
- Preserve Order: The final reasoning steps must follow the original sequence of the text.\\
- Completeness: All text from the original thinking trace must be included in the 'content' of the output steps. Do not omit any information.\\
- Informative Titles: Titles should be concise and describe the function of the step.\\
- Good Title Examples: "Deconstruct Problem", "Recall Known Facts", "Identify Key Constraints", "Formulate a Plan", "Analyze Clue 1", "Execute Calculation", "Resolve Contradiction", "Evaluate Hypothesis", "Synthesize Final Answer".\\
- Avoid Vague Titles: Do not use titles like "Step 1" or "Thinking".\\

Now, perform the task on the following input.
\begin{verbatim}
{thinking trace here}
\end{verbatim}
\end{tcolorbox}
\endgroup

\subsection{Meta-Summarization Operator}
\label{prompt:meta-summarize}

This prompt groups step summaries into canonical reasoning-function categories.

\phantomsection\label{prompt:box-meta-summarize}
\begingroup
\begin{tcolorbox}[promptbox, title=Prompt for the Meta-Summarization Operator]
ROLE:\\
You are an expert AI assistant specializing in the meta-analysis of reasoning processes. Your primary skill is to identify the underlying function of a reasoning step and categorize it according to a canonical, abstract model of thought.\\

TASK:\\
You will be given a list of titles, each summarizing a single step from a reasoning process. Your task is to cluster these titles based on their fundamental reasoning function. The goal is to produce a small set of categories that represent the core building blocks of the reasoning trace.\\

INPUT FORMAT:\\
A list of reasoning step titles, one per line. These titles may be in various languages.\\

OUTPUT FORMAT:\\
Your output MUST be a single, valid JSON object. The keys of the object must be cluster identifiers ("c0", "c1", "c2", etc.). Each value must be an object with two keys:\\
- "topic": A canonical, English topic name for the cluster.\\
- "data": A list containing all the original title strings belonging to that cluster.\\

Do not include any text or markdown formatting before or after the JSON object.\\

Example Structure:\\
\{
 "c0": \{
     "topic": "Canonical Topic 1",
     "data": ["title a", "title b", ...]
 \},
 "c1": \{
     "topic": "Canonical Topic 2",
     "data": ["title c", "title d", ...]
 \},
 ...
\}\\

CRITERIA FOR CLUSTERING AND NAMING:

1. Functional Clustering: The primary criterion for grouping titles is their purpose or function within the reasoning process. Group titles that describe the same type of logical step or cognitive action.

2. Canonical Topic Naming: The "topic" for each cluster must be a normalized, high-level label for that function.
   - It must be in clear and concise English.
   - It should be action-oriented where possible (e.g., using gerunds like "Identifying," "Calculating," "Verifying").
   - Good Topic Examples: "Problem Setup", "Problem Decomposition", "Information Extraction", "Formulating a Plan", "Uncertainty Management", "Self Checking", "Verifying Solution", "Synthesizing Final Result".

3. Parsimony: Be parsimonious. Aim for a limited, meaningful number of clusters. If two titles represent very similar functions (e.g., "Analyze the Original Sentence and Identify Key Phrases" and "Understand the Task and Initial Observations"), they belong in the same cluster. Merge similar functions under a single, more abstract topic.

4. Exhaustiveness: Every single title from the input list must be assigned to exactly one cluster. No titles should be left out.

5. Multilingual Semantics: The clustering must be based on semantic meaning, transcending language barriers. For example, a title in Spanish like "Calcular el costo total" should be in the same cluster as the English "Calculate the total cost."\\

Now, perform the task on the following input titles.
\begin{verbatim}
{summaries here}
\end{verbatim}
\end{tcolorbox}
\endgroup

\subsection{Reasoning Pattern Labeling}\label{prompt:label}

This prompt assigns each step summary to a meta-summary component.

\phantomsection\label{prompt:box-labeling}
\begingroup
\begin{tcolorbox}[promptbox, title=Prompt for Reasoning Pattern Labeling]
ROLE:\\
You are a meticulous text categorization expert. Your task is to analyze a "Summary" of a single reasoning step and classify it into one of the components of a "Meta-Summary". You will be given the summary itself, the full reasoning step it was derived from, and the entire thinking trace for complete context.\\

GOAL:\\
Determine which component of the "Meta-Summary" the "Summary" directly contributes to or is a part of.\\

INPUTS:\\
- "Meta-Summary": The final, high-level summary of the entire thinking process, broken down into numbered components (c0, c1, c2, etc.). These are your target categories.\\
- "Summary": The short text you must categorize.\\
- "Reasoning Step": The full text of the step that was summarized. Use this as the primary context to understand the "Summary".\\
- "Original Thinking Trace": The complete, step-by-step thought process. Use this for broader context if the "Reasoning Step" is ambiguous.\\

INSTRUCTIONS:\\
1.  First, carefully read the "Meta-Summary" to understand all the available categories (c0, c1, etc.).\\
2.  Next, read the "Summary" and its corresponding "Reasoning Step".\\
3.  Compare the "Summary" (clarified by the "Reasoning Step") to each component of the "Meta-Summary".\\
4.  Identify the single best component that the "Summary" logically fits into.\\
5.  If the "Summary" is a general procedural statement (e.g., "Now I will begin"), redundant meta-commentary, irrelevant, or does not logically contribute to any specific component, assign it the label "None".\\

OUTPUT REQUIREMENTS:\\
You MUST return ONLY the single component label (e.g., "c0", "c1", "c2") or the word "None". Do not include explanations, apologies, or any other text. The output must be a single, clean label.\\

DATA TO ANALYZE:

Meta-Summary:
\begin{verbatim}
{meta summary here}
\end{verbatim}

Summary to Categorize:
\begin{verbatim}
{summary here}
\end{verbatim}

Corresponding Reasoning Step:
\begin{verbatim}
{reasoning step here}
\end{verbatim}

Original Thinking Trace:
\begin{verbatim}
{thinking trace here}
\end{verbatim}
\end{tcolorbox}
\endgroup

\clearpage
\twocolumn
\section{Details of K-Means Clustering}\label{app:kmeans-details}

This section describes our data-guided procedure for selecting the number of clusters. For each of the \(K=128\) subsamples, the LLM independently induces a variable-sized set of candidate patterns. We use the modal number of patterns across subsamples as an initial estimate, \(\hat{L}\), and compare consolidation solutions with \(\hat{L}-1\), \(\hat{L}\), and \(\hat{L}+1\) clusters. We also consider larger values when they recover additional nonredundant functions. We select the most parsimonious solution that preserves the semantic coverage of the pooled candidates without introducing substantial redundancy. This broader search is necessary because each subsample captures only part of the full reasoning structure and may therefore undersegment it.

\section{HMS-Induced Reasoning Patterns}
\label{app:hms-patterns}

\subsection{Domain-General Pattern Definitions}
\label{app:l0-patterns}

We define the Level-0 functions used to interpret HMS clusters as follows. These labels denote functional categories rather than a prescribed generation pipeline; in individual traces, the corresponding reasoning steps may interleave or recur.

\noindent \textbf{(1) Understanding and Planning (U/P):} Interpreting the source text and task context, identifying relevant linguistic, cultural, or stylistic constraints, and formulating an initial translation strategy.

\noindent \textbf{(2) Translating and Drafting (T/D):} Producing candidate target-language renderings, including phrase-, clause-, and sentence-level translation decisions.

\noindent \textbf{(3) Refining and Verifying (R/V):} Checking the draft for fidelity, terminology, grammar, fluency, and style, and revising it as needed.

\subsection{HMS-Induced Patterns Across Datasets}
\label{app:qwen32b-meta-summaries}

Table~\ref{tab:hms-patterns} reports the HMS-induced pattern labels for Qwen-32B across three datasets. Level-0 labels summarize the dominant functional roles of reasoning steps, while Level-1 labels describe finer-grained sub-patterns within each Level-0 group.

\begin{table*}[tp!]
\centering
\small
\setlength{\tabcolsep}{0.01pt}
\renewcommand{\arraystretch}{1.08}
\resizebox{\textwidth}{!}{%
\begin{tabular}{p{0.10\textwidth}p{0.10\textwidth}p{0.3\textwidth}p{0.5\textwidth}}
\toprule
\textbf{Dataset} & \textbf{Stage} & \textbf{Level-0 Pattern} & \textbf{Level-1 Sub-patterns} \\
\midrule
\multirow{17}{*}{WMT}
  & \multirow{5}{*}{\textbf{U/P}}
  & \multirow{5}{*}{Understanding \& Planning}
  & Understanding Task and Context \\
  &&& Identifying Linguistic \& Cultural Nuances \\
  &&& Understanding User Intent and Context \\
  &&& Formulating \& Planning Translation Strategy \\
  &&& Source Text Analysis and Deconstruction \\
\cmidrule{2-4}
  & \multirow{6}{*}{\textbf{T/D}}
  & \multirow{6}{*}{Translating \& Drafting}
  & Initial Translation and Drafting \\
  &&& Sentence-Level Translation and Analysis \\
  &&& Segmented Translation and Analysis \\
  &&& Translating Specific Phrases and Terms \\
  &&& Handling Specific Terminology and Nuances \\
  &&& Refinement and Finalization of Translation \\
\cmidrule{2-4}
  & \multirow{6}{*}{\textbf{R/V}}
  & \multirow{6}{*}{Refining \& Verifying}
  & Accuracy and Fidelity Verification \\
  &&& Terminology and Word Choice Refinement \\
  &&& Grammar and Syntax Refinement \\
  &&& Contextual and Cultural Nuance Evaluation \\
  &&& Fluency and Naturalness Refinement \\
  &&& Overall Quality Assurance and Finalization \\
\midrule
\multirow{10}{*}{CMT}
  & \multirow{3}{*}{\textbf{U/P}}
  & \multirow{3}{*}{Understanding \& Planning}
  & Source Text Analysis \& Key Element Identification \\
  &&& Translation Strategy Formulation \\
  &&& Initial Task and Context Understanding \\
\cmidrule{2-4}
  & \multirow{3}{*}{\textbf{T/D}}
  & \multirow{3}{*}{Translating Specific Elements}
  & Translating Sentence Segments and Clauses \\
  &&& Translating Proper Nouns and Specific Entities \\
  &&& Handling Grammatical Structures \& Contextual Nuances \\
\cmidrule{2-4}
  & \multirow{4}{*}{\textbf{R/V}}
  & \multirow{4}{*}{Refining \& Verifying}
  & Verification and Accuracy Check \\
  &&& Grammar and Syntax Refinement \\
  &&& Overall Translation Synthesis and Finalization \\
  &&& Terminology and Proper Noun Verification \\
\midrule
\multirow{17}{*}{DRT}
  & \multirow{4}{*}{\textbf{U/P}$_1$}
  & \multirow{4}{*}{Initial Understanding \& Analysis}
  & Contextual and User Intent Analysis \\
  &&& Detailed Semantic and Lexical Analysis \\
  &&& Synthesizing and Interpreting Overall Meaning \\
  &&& Initial Task and Context Understanding \\
\cmidrule{2-4}
  & \multirow{4}{*}{\textbf{U/P}$_2$}
  & \multirow{4}{*}{Translation Strategy \& Planning}
  & Ensuring Fidelity to Tone, Style, and Poetic Elements \\
  &&& Establishing Overall Translation Goals and Principles \\
  &&& Analyzing User Intent and Context \\
  &&& Addressing Specific Linguistic and Stylistic Elements \\
\cmidrule{2-4}
  & \multirow{5}{*}{\textbf{T/D}}
  & \multirow{5}{*}{Translating \& Drafting}
  & Figurative Language and Idiom Translation \\
  &&& Direct Phrase and Clause Translation \\
  &&& Contextual and Nuance-Based Translation \\
  &&& Translation Strategy and Refinement \\
  &&& Segmented Translation and Analysis \\
\cmidrule{2-4}
  & \multirow{4}{*}{\textbf{R/V}}
  & \multirow{4}{*}{Refinement \& Verification}
  & Overall Translation Review and Refinement \\
  &&& Verifying Accuracy and Fidelity \\
  &&& Specific Terminology and Phrase Refinement \\
  &&& Refining Fluency and Naturalness \\
\bottomrule
\end{tabular}%
}
\caption{HMS-derived hierarchical reasoning pattern labels for Qwen-32B across three translation datasets. For DRT, HMS separates initial interpretation from translation-strategy planning, reflecting the stronger semantic and stylistic demands of literary translation.}
\label{tab:hms-patterns}
\end{table*}
\section{Domain-Specific Allocation at Level-1}
\label{appendix:domain-specific}

\begin{figure*}[ht]
    \centering

    \begin{subfigure}{\textwidth}
        \centering
        \includegraphics[width=\textwidth]{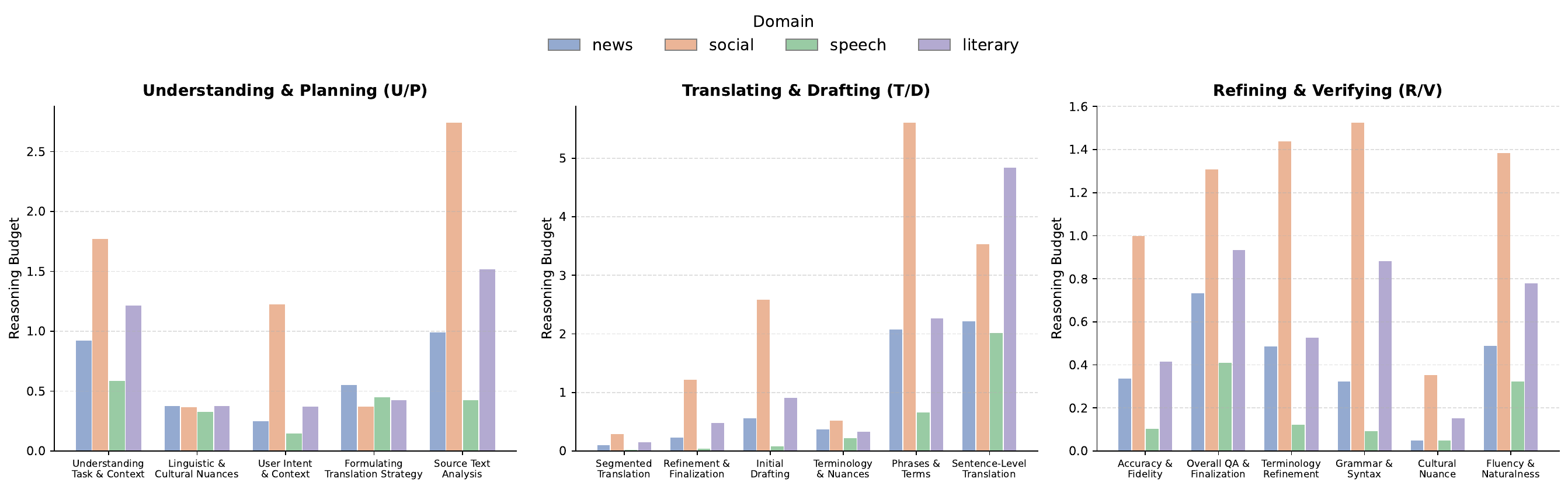}
        \caption{Reasoning budget: absolute allocation of Level-1 reasoning patterns}
        \label{fig:WMT24-L1-Distribution-Budget}
    \end{subfigure}


    \begin{subfigure}{\textwidth}
        \centering
        \includegraphics[width=\textwidth]{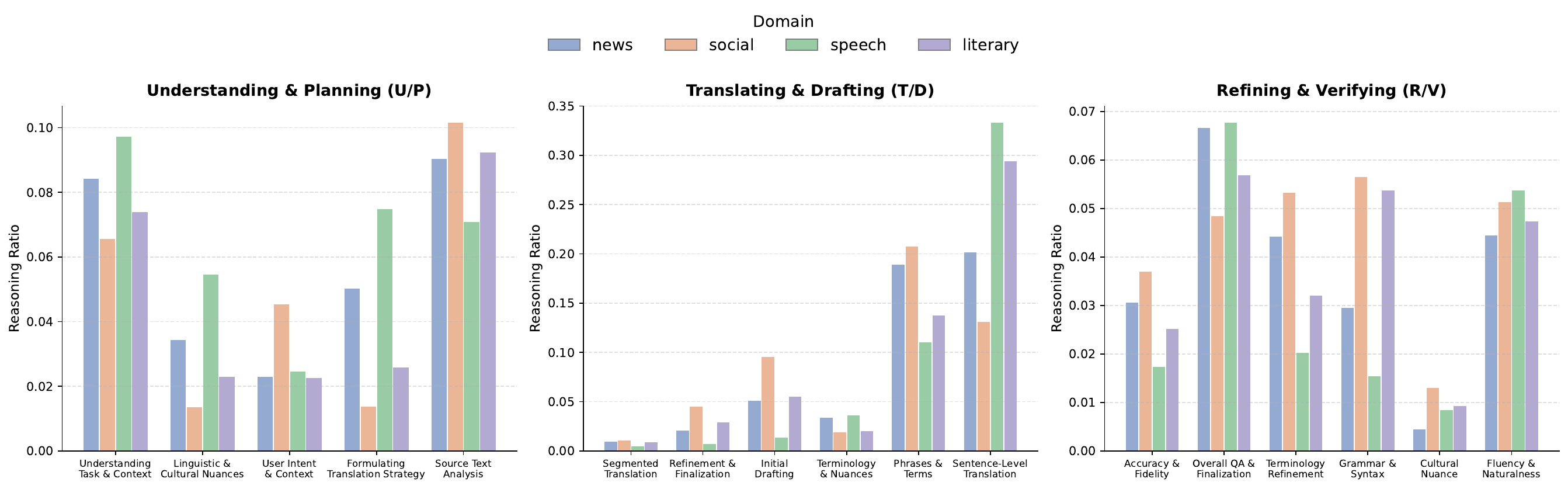}
        \caption{Reasoning ratio: relative allocation of Level-1 reasoning patterns}
        \label{fig:WMT24-L1-Distribution-Ratio}
    \end{subfigure}

    \caption{Domain-specific allocation of WMT24 Level-1 reasoning patterns for Qwen-32B. Absolute budgets and relative ratios show how reasoning effort is distributed across fine-grained patterns within each functional stage.}
    \label{fig:WMT24-L1-Distribution}
\end{figure*}

To complement the Level-0 analysis, we examine how Qwen-32B allocates reasoning across Level-1 patterns within each functional stage on WMT24. Figure~\ref{fig:WMT24-L1-Distribution} reports both Reasoning Budget and Reasoning Ratio: the former measures the token budget assigned to each pattern, while the latter normalizes by total reasoning length to support cross-domain comparison.

\paragraph{Understanding and Planning (U/P).} 
U/P reasoning is anchored in task understanding and source-text analysis, but its sub-pattern mix varies by domain. Speech allocates relatively more effort to task/context understanding, linguistic and cultural nuances, and strategy formulation, whereas social text places the most emphasis on source-text analysis and user-intent/context reasoning. These shifts show that the same U/P stage foregrounds different preparatory cues across domains.

\paragraph{Translating and Drafting (T/D).} 
T/D shows strong domain-specific variation. Speech and literary texts allocate the largest relative shares to sentence-level translation, while news and social text place more emphasis on phrase- and term-level translation. Social text also has the largest initial-drafting share and the highest absolute budgets across many T/D sub-patterns, reflecting both relative preferences and longer traces.

\paragraph{Refining and Verifying (R/V).} 
R/V reasoning is spread across multiple revision functions. Overall QA/finalization is prominent for news and speech, fluency and naturalness remains substantial across domains, and social text assigns relatively more effort to grammar, terminology, and accuracy checks. Thus, post-drafting reasoning adapts to domain-specific revision needs rather than acting as a uniform verification step.

Overall, the Level-0 structure provides a stable domain-general organization, while Level-1 allocations reveal domain-specific adaptations in reasoning effort.
\section{Reasoning Pattern Effectiveness}
\label{app:reasoning-pattern-effectiveness}

\begin{figure*}[tp!]
    \centering
    \includegraphics[width=1.0\textwidth]{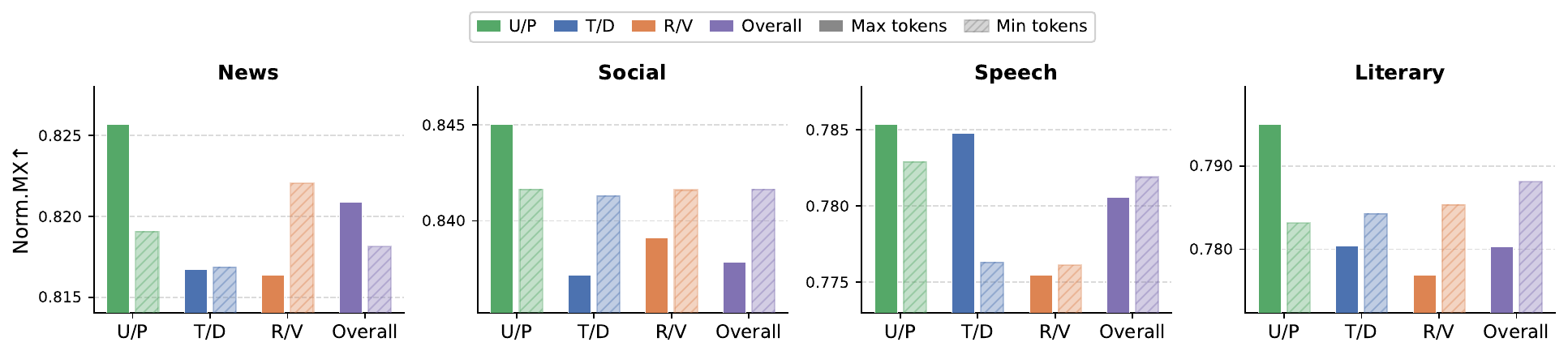}
    \caption{Qwen-32B normalized reference-based MetricX scores under maximum versus minimum reasoning lengths on WMT24. Longer U/P is the most stable positive signal, while T/D and R/V effects vary more by domain.}
    \label{fig:wmt24_l1_qwen32b_metricx_rb}
\end{figure*}

\begin{figure*}[tp!]
    \centering
    \includegraphics[width=1.0\textwidth]{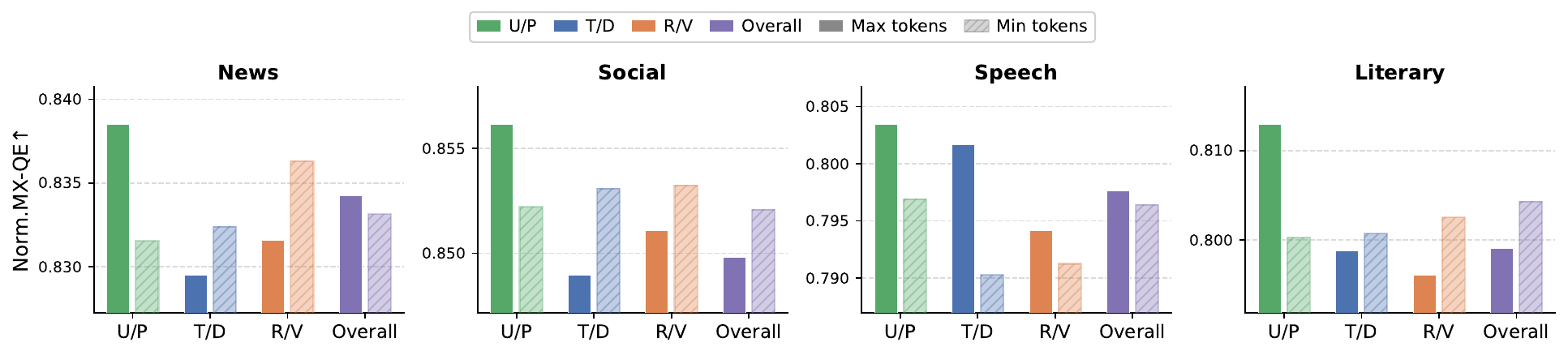}
    \caption{Qwen-32B normalized MetricX-QE scores under maximum versus minimum reasoning lengths on WMT24. Longer U/P is generally favorable; longer T/D and R/V show more selective, domain-dependent patterns.}
    \label{fig:wmt24_l1_qwen32b_metricx_qe}
\end{figure*}

\begin{figure*}[tp!]
    \centering
    \includegraphics[width=1.0\textwidth]{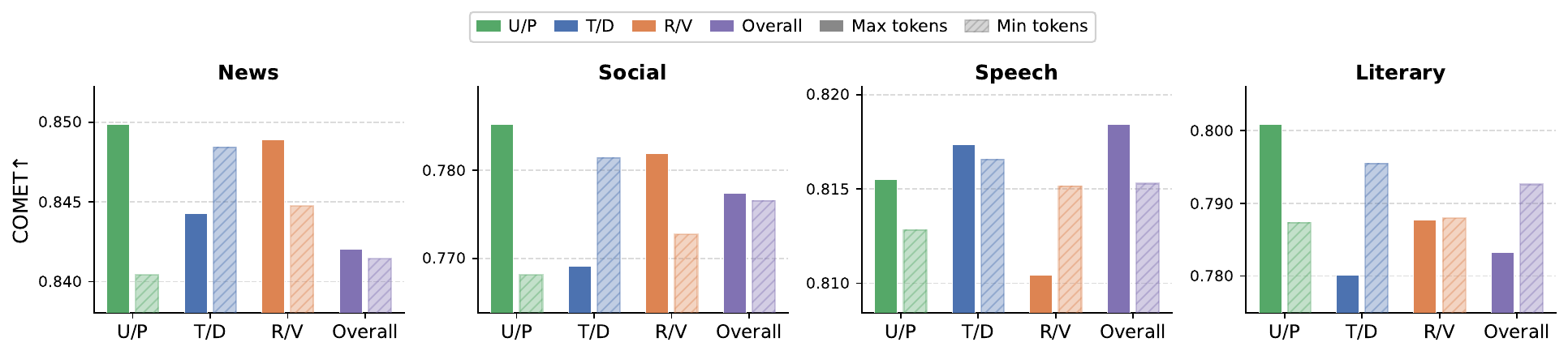}
    \caption{Llama-8B COMET scores under maximum versus minimum reasoning lengths on WMT24. The comparison broadly follows the Qwen-32B trend, with U/P more stable than T/D or R/V across domains.}
    \label{fig:wmt24_l1_llama8b_comet}
\end{figure*}

\begin{figure*}[tp!]
    \centering
    \includegraphics[width=1.0\textwidth]{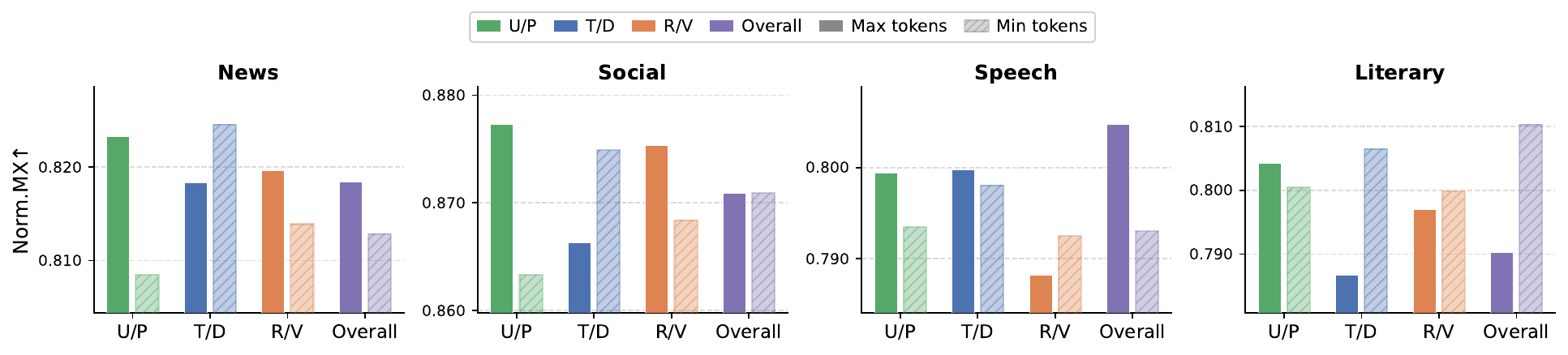}
    \caption{Llama-8B normalized reference-based MetricX scores under maximum versus minimum reasoning lengths on WMT24. U/P shows the clearest positive pattern, whereas T/D and R/V depend more on domain.}
    \label{fig:wmt24_l1_llama8b_metricx_rb}
\end{figure*}

\begin{figure*}[tp!]
    \centering
    \includegraphics[width=1.0\textwidth]{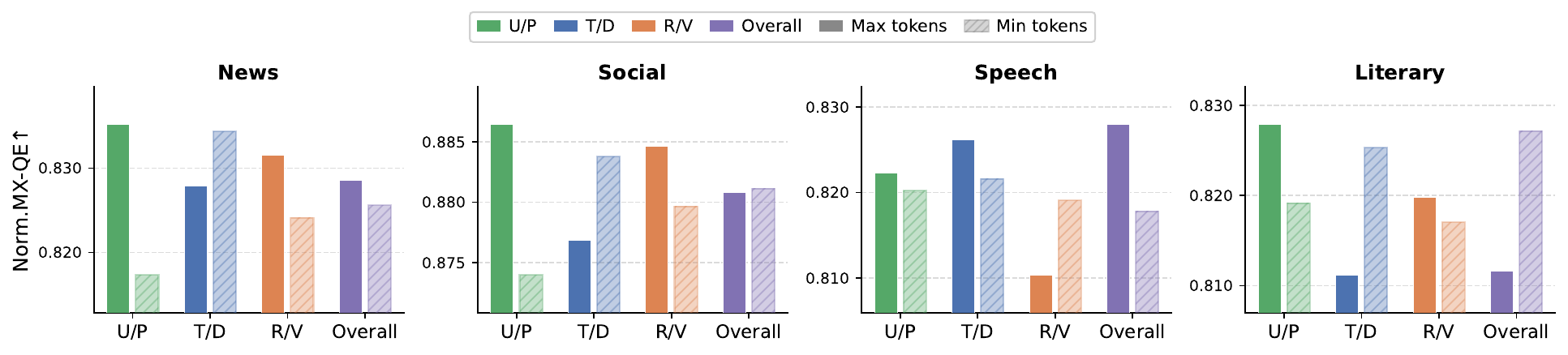}
    \caption{Llama-8B normalized MetricX-QE scores under maximum versus minimum reasoning lengths on WMT24. Longer reasoning is not uniformly beneficial: U/P is comparatively robust, while T/D and R/V are more conditional.}
    \label{fig:wmt24_l1_llama8b_metricx_qe}
\end{figure*}

This section reports the supplementary WMT24 comparisons used to examine how Level-1 reasoning length relates to translation quality. The main text shows Qwen-32B COMET results; Figures~\ref{fig:wmt24_l1_qwen32b_metricx_rb}--\ref{fig:wmt24_l1_llama8b_metricx_qe} add normalized reference-based MetricX, normalized MetricX-QE, and Llama-8B settings. Each figure compares outputs with the minimum and maximum observed reasoning lengths for a functional stage, so the results should be interpreted as stage-specific associations rather than causal interventions.

Overall, the supplementary results are consistent with the main trend but not uniform in every metric--model pair. Longer U/P is the most stable positive signal, indicating that additional source analysis and planning often correlate with better quality. Longer T/D is more selective and is most often beneficial for speech, where informal or disfluent input may require local reformulation. R/V varies most by domain and metric, suggesting that extra revision helps only when it addresses real adequacy or fluency issues and can otherwise lead to unnecessary changes.

These results support pattern-aware allocation of reasoning length: planning is generally the safest place to add budget, drafting should be expanded selectively, and verification requires domain- and metric-sensitive calibration.

\end{document}